\documentclass{article}

\usepackage{microtype}
\usepackage{graphicx}
\usepackage{subfigure}
\usepackage{booktabs} 

\usepackage[accepted]{icml2023}

\usepackage{amsmath,amsfonts,bm}
\usepackage{subfigure}
\definecolor{myblue}{rgb}{0.00, 0.45, 0.70}
\definecolor{mygreen}{rgb}{0.01, 0.62, 0.45}
\definecolor{myyellow}{rgb}{0.9, 0.5, 0}
\definecolor{mygrey}{rgb}{0.5, 0.5, 0.5}
\definecolor{myred}{rgb}{0.84, 0.37, 0.00}

\newcommand{\Grad}{{\mathcal{G}}}

\newcommand{\Wt}{{ \mathcal{W}^{(t)}}}

\newcommand{\unif}{\textsc{unif}}

\def\1{\bm{1}}

\newcommand{\RNum}[1]{\uppercase\expandafter{\romannumeral #1\relax}}

\def\vtheta{{\bm{\theta}}}
\def\valpha{{\bm{\alpha}}}
\def\vepsilon{{\bm{\varepsilon}}}

\def\vw{{\bm{w}}}

\DeclareMathAlphabet{\mathsfit}{\encodingdefault}{\sfdefault}{m}{sl}
\SetMathAlphabet{\mathsfit}{bold}{\encodingdefault}{\sfdefault}{bx}{n}

\def\gW{{\mathcal{W}}}

\newcommand{\R}{\mathbb{R}}

\DeclareMathOperator*{\argmin}{arg\,min}

\usepackage{amsmath}
\usepackage{amssymb}
\usepackage{mathtools}
\usepackage{amsthm}
\usepackage{silence}
\usepackage[capitalize,noabbrev]{cleveref}

\theoremstyle{plain}

\theoremstyle{definition}

\theoremstyle{remark}

\usepackage[textsize=tiny]{todonotes}
\usepackage{url}

\usepackage{adjustbox}

\definecolor{DarkGreen}{rgb}{0.1,0.5,0.1}
\definecolor{DarkRed}{rgb}{0.5,0.1,0.1}
\definecolor{DarkBlue}{rgb}{0.1,0.1,0.7}
\definecolor{DarkYellow}{rgb}{.79,.79,0}

\usepackage{latexsym}
\usepackage{tabularx}
\usepackage{graphicx}
\usepackage{multirow}
\usepackage{booktabs}
\usepackage{enumitem}
\usepackage{amsmath}
\usepackage{pifont}
\usepackage{amssymb}
\usepackage{color, colortbl}
\usepackage{url}
\usepackage{float}
\usepackage{appendix}
\usepackage{bbold}
\usepackage{xspace}
\usepackage{adjustbox}	
\usepackage[T1]{fontenc}
\usepackage{changepage}  

\usepackage{algorithm}
\usepackage{algpseudocode}

\newcommand{\doge}{{\textsc{DoGE}}\xspace}
\newcommand\dogeemoji{\raisebox{-3.5pt}{\includegraphics[width=1.3em]{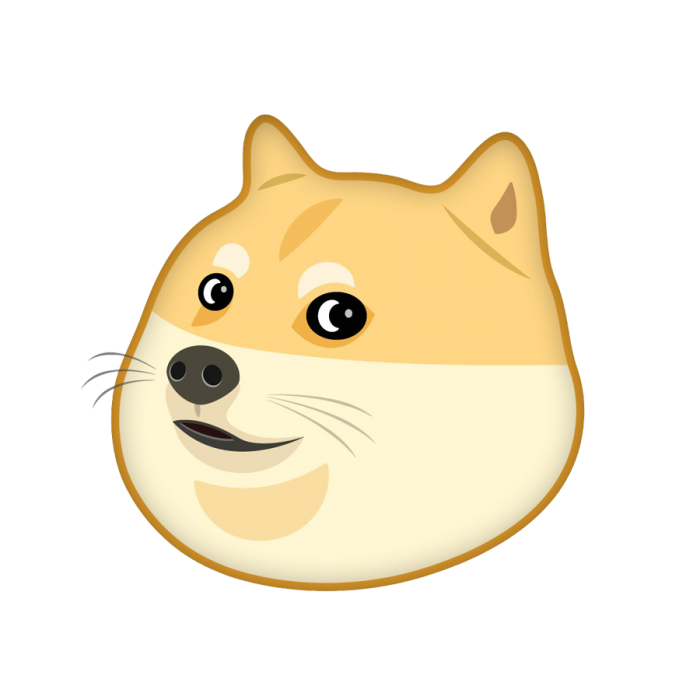}}}

\newcommand\ignore[1]{}
\icmltitlerunning{\textsc{DoGE}: Domain Reweighting with Generalization Estimation}

\begin{document}

\twocolumn[
\icmltitle{\textsc{DoGE}\dogeemoji{}: Domain Reweighting with Generalization Estimation}

\begin{icmlauthorlist}
\icmlauthor{Simin Fan}{epfl}
\icmlauthor{Matteo Pagliardini}{epfl}
\icmlauthor{Martin Jaggi}{epfl}
\end{icmlauthorlist}

\icmlaffiliation{epfl}{EPFL, Switzerland}
\icmlcorrespondingauthor{Simin Fan}{simin.fan@epfl.ch}
\icmlkeywords{Language Models, Pretraining, Data}
\vskip 0.3in]


\printAffiliationsAndNotice{}  

\begin{abstract}
The coverage and composition of the pretraining data significantly impacts the generalization ability of Large Language Models (LLMs). Despite its importance, recent LLMs still rely on heuristics and trial and error to increase or reduce the influence of data-domains. We propose \textsc{DO}main reweighting with \textsc{G}eneralization \textsc{E}stimation (\textsc{DoGE}), which optimizes the probability of sampling from each domain (domain weights) in a principled way. Our approach is a two-stage process consisting of (i) training a \emph{proxy model} to obtain domain weights using a bi-level optimization algorithm; (ii) training a larger \emph{base model} by sampling training domains according to the learned domain weights. In our experiments, we extensively show how \textsc{DoGE} improves the generalization of the base model to any target data mixture. 
On the SlimPajama dataset, our base model gets better perplexity and few-shot reasoning accuracies across $6$ tasks compared to baseline methods. Moreover, aiming to generalize to out-of-domain target tasks, which is unseen in the pretraining corpus (OOD domain), \textsc{DoGE} can effectively identify inter-domain dependencies, and consistently achieves better test perplexity on the target domain. 
\end{abstract}
\vspace{-2em}
\section{Introduction}
\label{sec:intro}
Pretrained Large Language Models (LLMs) demonstrate impressive generalization abilities, making them the workhorse of today's NLP research and many practical use cases~\citep{devlin2019bert, brown2020gpt3,chowdhery2022palm, touvron2023llama, touvron2023llama2}. They are trained on very large text corpora collected from various source domains to obtain a generalization ability, which enables an efficient adaptation to specific downstream tasks by fine-tuning. 
The composition of the pretraining corpus often depend on the accessibility of each data sources. 
For example, $72.6\%$ of RedPajama~\citep{together2023redpajama} are sampled from CommonCrawl, while only $1.7\%$ from Stackexchange. 
While recent research has demonstrated the significance of the quantity and quality of the pretraining corpus~\citep{kaplan2020scaling, hoffmann2022chinchila, longpre2023pretrainers}, there are few explorations into how its composition from various source domains could contribute to the generalization ability of the language model~\citep{lee2023scale, Hashimoto2021ModelPS, xie2023doremi}. The \emph{domain weights} adopted by current state-of-the-art LLMs are mostly determined by heuristics~\citep{gao2020pile} or tuned according to a series of downstream tasks~\citep{du2022glam}, which can be sub-optimal and costly.

Recently, \citet{xie2023doremi} proposed a learnability-based domain reweighting framework \textsc{DoReMi}, which settles \emph{domain weights} using two small-scale auxiliary models: first, a reference model is "well-trained" using uniform domain weights; next, a second auxiliary model---referred to as proxy model---is trained from scratch with the objective to find domain weights that minimize the worst-case \emph{excess loss}, i.e. the per-domain loss gap between the proxy model and the well-trained reference model. The \emph{excess loss} is interpreted as an estimation for the remaining learnability of a given domain at each training step---a large gap indicating the proxy model can further learn to model the associated domain. Despite the encouraging empirical results of \textsc{DoReMi}, minimizing the worst-case loss gap (i) creates a strong dependency on the well-trained model whose capacity can strongly influence the overall accuracy and requires appropriate tuning, and (ii) creates a dissonance between the ideal goal of minimizing the average validation loss across domains and the employed objective which seeks to simply mimic the well-trained model. Moreover, this approach cannot be used when the target domains are different from the training domains.

To mitigate these issues, we propose \textsc{Do}main reweighting with \textsc{G}eneralization \textsc{E}stimation (\doge), which finds optimal \emph{domain weight} distributions by explicitly optimizing for best generalization to a given set of domains. 
We follow the two-stage process of \textsc{DoReMi} which consists of first obtaining optimized domain weights by training a small-scale proxy model, and, in the second stage, training a final larger model on data sampled according to those weights. In contrast to \textsc{DoReMi}, \textsc{DoGE} only requires the training of one proxy model. Moreover, we found \textsc{DoGE} to be less dependent on the capacity of this proxy model (see \S~\ref{apdx:robustness}). 
When training the proxy model, at timestep $t$, we re-weight the gradient from each source domain to greedily minimize the average target domain loss at the next step $t+1$. Our derivation in \S~\ref{sec:method} shows that the resulting algorithm up-weights training domains with a large gradient alignment (inner-product) with the target domains, which reflects the principle:

\begin{adjustwidth}{2em}{2em}
\emph{A data domain should receive a large weight if it contributes to the learning of target domains.}
\end{adjustwidth}

\begin{figure}[htbp!]
    \centering
    \includegraphics[width=0.5\textwidth]{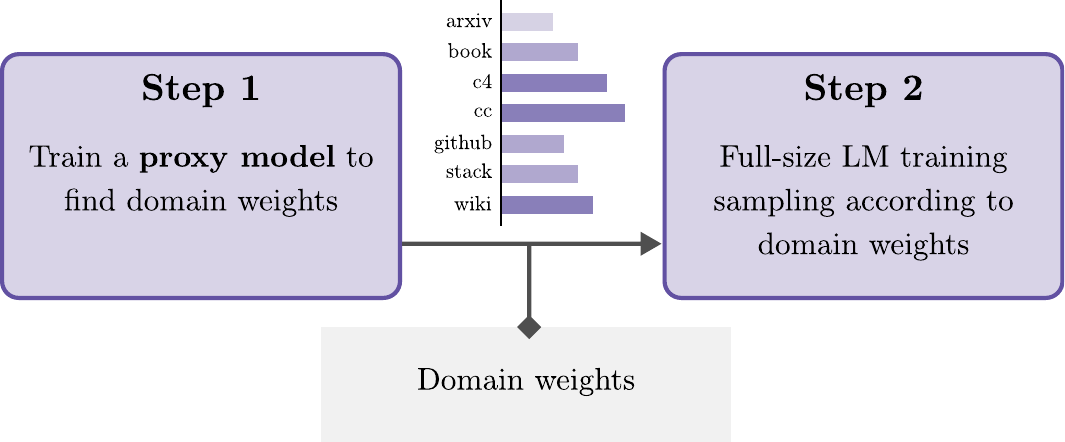}
    \vspace{-0.5em}
    \caption{\textbf{Summary of \doge\dogeemoji.} Our method consists of two steps. In \textbf{Step 1}, we learn domain weights which maximize the generalization of the proxy model to the target domains. The resulting domain weights are then used in \textbf{Step 2} to train a---potentially larger---language model. 
    }
    \vspace{-1em}
    \label{fig:overview}
\end{figure}

Similarly to \textsc{DoReMi}, the final domain weights are obtained by averaging the domain weights over the training of the proxy model. The base model is then trained by sampling its training data according to the final domain weights. A visual overview of the \doge method is shown in Fig.~\ref{fig:overview}. 

\textbf{Contributions.} We summarize our contributions as follows:
\begin{itemize}
    \item We introduce and rigorously derive \doge, an efficient and effective domain reweighting framework, which explicitly aims to generalize to a specific set of target domains (\S~\ref{sec:method});
    \item We empirically show that our method outperforms strong baselines including \textsc{DoReMi} in terms of (i) average perplexity, and (ii) few-shot reasoning capabilities across 6 tasks (\S~\ref{sec:universal-exp});
    \item We show how \doge can handle cases where the target domains are different from the training domains, and consistently outperforms the baseline with uniform domain weights (\S~\ref{sec:ood-exp}). 
\end{itemize}
\vspace{-1em}

\section{Domain Reweighting with Generalization Estimation}\label{sec:method}

In this section, we motivate and derive \doge, for the goal of re-weighting training domains $D_{train}\triangleq \{D_1, \ldots, D_k\}$ to improve the model's generalization to a given set of target domains. 
We distinguish two scenarios for generalization: (1) \textbf{Universal generalization}, where the target objective is to minimize the validation loss across all source domains $\{D_1, \ldots, D_k\}$; as well as (2) \textbf{Out-of-domain generalization} where we aim at minimizing the validation loss on a specific target domain ($D_{ood}$), while $D_{ood} \notin D_{train}$. The first case applies in most of the scenarios for LLM pretraining, where no specific downstream target has been set. The later case is especially relevant when considering generalization to specific target domain datasets (e.g. science, low-resource languages) which are too small to have a significant impact when used during pretraining. 

\textbf{Setup \& notation.}
Let $D_{train}\triangleq \{D_1, \ldots, D_k\}$ be a large corpus split into $k$ domains according to meta-attributes (e.g. source, topic). We aim to find \textbf{\emph{domain weights}} over the probability simplex $\valpha \in \Delta^k \subset\R^k$. The final data mixture used to train the full-size language model is constructed by first sampling a domain according to the domain-wise distribution $\valpha$, followed by uniformly sampling a batch $B$ from that domain ($B \sim \unif(D_i)$). Overall, this leads to the instance-wise distribution $P_\valpha \triangleq  \sum_{i=1}^k \alpha_i \cdot \unif(D_i)$. 
In the following, we will describe how to optimize $\valpha$ guided by training a proxy model of parameters $\vtheta$ on $D_{train}$. 
We denote by $l_i(\vtheta)$ the expected next token prediction loss of the proxy model on domain $D_i$. 
Let $\Bar{l}(\vtheta)\triangleq \frac{1}{k}\sum_{i \in [k]} l_i(\vtheta)$ be the average loss across all $k$ domains. Let $|D|$ refer to the number of samples in $D$.

\textbf{Universal generalization.} In the case of universal generalization, our goal is to minimize $\Bar{l}(\vtheta)$. This posit that all $k$ given training domains have the same importance. As a point of comparison, note that the classical loss used to train large language models is $\displaystyle{\tilde{l}(\vtheta) = \sum_{i \in [k]}\frac{|D_i|}{|D_{train}|}l_i(\vtheta)}$ which could severely bias to domains with larger scale. One naive approach could consist in re-weighting samples by the inverse of the sampling probability: $\displaystyle{\tilde{l}(\vtheta) = \sum_{i \in [k]}\tilde{\alpha}_i\frac{|D_i|}{|D_{train}|}l_i(\vtheta)}$ with $\displaystyle{\tilde{\alpha}_i=\frac{|D_{train}|}{|D_i|}}$, however, this approach ignores everything of the complex intra-domain interactions considering the nature of the textual corpus which (i) have inevitable lexical, syntactic or semantic overlaps, and (ii) can be more or less challenging to learn. In practice, this naive uniform sampling approach provides a strong baseline but often hinders the generalization compared to other methods (see \S~\ref{sec:experiment}). 


We instead propose to optimize domain weights $\valpha\in\Delta^k$ along the training of the proxy model $\vtheta$, as a stochastic bi-level optimization problem: \vspace{-2mm}
\begin{equation*}
\begin{aligned}
    \valpha &\in \argmin_{\valpha\in\Delta^k} \sum_{i \in [k]}l_i(\vtheta^\star(\valpha))\\
    s.t.\ \vtheta^\star(\valpha) &\in \argmin_\vtheta \qquad\! \sum_{i \in [k]}\alpha_i l_i(\vtheta)
\end{aligned} 
\end{equation*}
In the inner loop~\eqref{equ:obj-t0}, the proxy model $\vtheta(\valpha)$ is updated using the rescaling factor $\valpha$; in the outer loop \eqref{equ:obj-t1}, we update~$\valpha$ to adapt to the target given the updated model status. To avoid complicated multi-step gradient unrolling, we only update~$\vtheta$ in the inner optimization problem over a single stochastic step:
\begin{equation}\label{equ:obj-t0}
\vtheta^{(t+1)} \triangleq \vtheta^{(t)} - \eta^{(t)} \sum_{i\in [k]}\alpha_i^{(t)} \nabla l_i(\vtheta^{(t)})
\end{equation}
where $\valpha^{(t)} \in \Delta^{k}$ is used to re-weight the loss from each domain at time-step~$t$, $\eta^{(t)}$ is the step size, and $\nabla l_i(\vtheta^{(t)})$ is a stochastic gradient for samples of $D_i$. 
The outer-loop in bi-level optimization techniques usually requires second-order derivatives \citep{grangier2023adaptive,zhou2023probabilistic}, which could introduce huge computation costs. Instead, we update $\valpha$ by a simpler fully first-order rule, which allows to reuse the gradients from the inner-loop.

Specifically, the update rule of the domain weights~$\valpha$ can be derived as follows. Denote $\Delta\vtheta^{(t)}=\vtheta^{(t+1)}-\vtheta^{(t)}$, at step~$t$, we aim to find the optimal $\valpha^{(t)}$ to minimize the original unweighted domain loss at the next step:
\begin{align}\label{equ:obj-t1}
\valpha_\star^{(t)} &= \argmin_{\valpha \in \Delta^{k}} \,\Bar{l}(\vtheta^{(t+1)}) \notag \\ 
&= \argmin_{\valpha \in \Delta^{k}} \sum_{i \in [k]} [l_i(\vtheta^{(t+1)}) - l_i(\vtheta^{(t)})] \\
&= \argmin_{\valpha \in \Delta^{k}} \sum_{i \in [k]} \langle \nabla l_i(\vtheta^{(t)}), \Delta\vtheta^{(t)}\rangle +o(\| \Delta\vtheta^{(t)} \|)\notag \\
&= \argmin_{\valpha \in \Delta^{k}} \sum_{i \in [k]} \!\!\big\langle \nabla l_i(\vtheta^{(t)}\!), -\eta^{(t)}\!\!\! \sum_{j\in [k]} \!\!\alpha_j \nabla l_j(\vtheta^{(t)}) \big\rangle+\vepsilon^{(t)},\notag
\end{align}
where $\vepsilon^{(t)}=o(\| \Delta\vtheta^{(t)} \|)=o(\| \valpha^{(t)} \|)$ \eqref{equ:obj-t0}, which is a high-order remainder in the Taylor expansion. 
Let $W^{(t)}_j \triangleq \langle\nabla l_j(\vtheta^{(t)}), \sum_{i \in [k]}\nabla l_i(\vtheta^{(t)}) \rangle$ be the stochastic \emph{generalization estimation function} on the $j^{th}$ domain. 
Intuitively, this quantity measures the alignment of the learning tasks across domains: a high $W^{(t)}_j$ means learning $D_j$ will also contribute to learning other domains. 
We write $\gW^{(t)}=[W^{(t)}_1, \ldots, W^{(t)}_k] \in\R^k$ for the vectorized generalization estimation scores across all domains.
We can rewrite the outer loop update \eqref{equ:obj-t1} simply as:\vspace{-1mm}
\begin{align}\label{equ:obj-t2}
\valpha_\star^{(t)} &= \argmin_{\valpha \in \Delta^{k}} -\eta^{(t)} \valpha^\top \gW^{(t)}+\vepsilon^{(t)}\ .
\end{align}
We solve \eqref{equ:obj-t2} by estimating $\vepsilon^{(t)}$ as the Bregman divergence $D_\Psi(\valpha \| \valpha^{(t-1)})$ with $\Psi(\valpha)=\sum_i \alpha_i \log(\alpha_i)$, which is a common technique in mirror descent \citep{NemYud83,BeckT03} :
\vspace{-0mm}
\begin{align}\label{equ:obj-t3}
 \valpha^{(t)} = \argmin_{\valpha \in \Delta^{k}} -\eta^{(t)} \valpha^\top {\gW}^{(t)} + \mu D_\Psi(\valpha \| \valpha^{(t-1)}),
\end{align}
with $\mu$ as a hyperparameter controls the strength of regularization. This yields the following multiplicative weights update rule, see e.g. \citep{BeckT03}:\vspace{-1mm}
\begin{align}\label{equ:update_dw}
 \valpha^{(t)} = \frac{\hat{\valpha}^{(t)}}{\sum_{i \in [k]} \hat{\alpha}_i^{(t)}}, 
\end{align}
with $\hat{\valpha}^{(t)} = \valpha^{(t-1)} \odot \exp\Big(\displaystyle{\frac{\eta^{(t)}{\gW}^{(t)}}{\mu}}\Big)$. 
We estimate the average domain loss $\sum_{i \in [k]}\nabla l_i(\vtheta^{(t)})$ by sampling another batch consisting of instances uniformly sampled from all domains. 
At each time-step $t$, we alternatively update $\valpha^{(t)}$ and $\vtheta^{(t)}$. The final algorithm is summarized in Alg.~\ref{alg:doge}. The detailed derivation is presented in Appendix (\S~\ref{apdx:algorithm}).

\textbf{Out-of-domain generalization.}
In the out-of-domain generalization scenario we want to generalize to a $D_{ood}$ domain that is not part of $D_{train}$. The above derivation still holds only with minor modifications: (i) we are now considering our objective to be $l_{ood}(\vtheta)$ instead of $\Bar{l}(\vtheta)$, and (ii) we now have $W^{(t)}_j \triangleq \langle \nabla l_j(\vtheta^{(t)}) ,\nabla l_{ood}(\vtheta^{(t)}) \rangle$, for clarity we call $\gW^{(t)}_{ood}=[W^{(t)}_1, \ldots, W^{(t)}_k]$. The update of $\valpha^{(t)}$ is the same as in \eqref{equ:update_dw} replacing $\gW^{(t)}$ with $\gW^{(t)}_{ood}$. The associated algorithm can be seen in App.~\ref{apdx:algorithm} (See Alg.~\ref{alg:doge-ood}), where all differences with universal generalization (Alg.~\ref{alg:doge}) are highlighted in blue. 

\textbf{Link between $\Wt$ and influence functions.}
Following~\cite{pruthi2020tracin}, given samples from a source and target domain $B_{s} \sim D_{s}$ and $B_{t} \sim D_{t}$, the influence of $D_{s}$ on $D_{t}$ can be estimated by $\mathcal{I}(B_{s}, B_{t})=\langle \nabla l_{s}(\vtheta), \nabla l_{t}(\vtheta) \rangle$. Considering the definition of $W^{(t)}_j$:
\begin{align}\label{equ:ge-decompose}
    W^{(t)}_j &\triangleq \langle\nabla l_j(\vtheta^{(t)}), \sum_{i \in [k]}\nabla l_i(\vtheta^{(t)}) \rangle \\ 
    &= \underbrace{\langle\nabla l_j(\vtheta^{(t)}), \sum_{i \in [k],i\neq j}\nabla l_i(\vtheta^{(t)})\rangle}_{\text{out-of-domain influence}} + \underbrace{\|\nabla l_j(\vtheta^{(t)})\|_2^2}_{\text{domain difficulty}} \notag
\end{align}
The first term in \eqref{equ:ge-decompose} estimates the sum of influences from all the other $k-1$ domains on the $j^{th}$ domain, while the second term denotes the magnitude of the gradient from domain~$D_j$. Intuitively, a domain should be up-weighted when (i) it contributes to the learning of other domains (high out-of-domain influence), or (ii)---in the universal generalization case---when the domain itself has not been learnt enough (high magnitude of gradient for this domain). Those two mechanisms are precisely what Equ.~\eqref{equ:obj-t2} expresses. 
\begin{algorithm}[ht!]
   \caption{\doge Domain Reweighting (for Universal Generalization).}
   \label{alg:doge}
\begin{algorithmic}[1]
   \State {\bfseries Input:} Domain data splits $D_1,\dots, D_k$, Proxy model weights $\vtheta^{(0)}$, Hyperparameters: number of training steps $T$, batch size $b$, step size $\eta^{(t)}$, Bregman coefficient $\mu$. 
   \State Initialize proxy weights $\vtheta^{(0)}$
   \State Initialize proxy domain weights $\valpha^{(0)} = \frac{1}{k}\mathbf{1}$
   \For{$t \in [T]$}
        \State \emph{Uniformly} sample batch $B^{(t)}=\{B_1^{(t)},\dots,B_k^{(t)}\}$ 
        \State Obtain $\nabla l_i(\vtheta^{(t)}, B_i^{(t)})$ for $i \in [k]$ \hfill 
        \State Compute $\gW^{(t)}$ 
        \State Update domain weights according to Eq.~\eqref{equ:update_dw}:\\
        $\qquad \qquad \displaystyle{\hat{\valpha}^{(t)} \gets \valpha^{(t-1)} \odot \exp(\eta^{(t)}\gW^{(t)}/\mu)}$ \\
        $\qquad \qquad \valpha^{(t)} \gets \displaystyle{\hat{\valpha}^{(t)}/\sum_{i=1}^k \hat{\alpha}^{(t)}_i}$
        \State Update $\vtheta^{(t)}$:\\
        $\qquad\   \vtheta^{(t+1)} = \vtheta^{(t)} - \eta^{(t)} \sum_{i\in [k]}\alpha_i^{(t)} \nabla l_i(\vtheta^{(t)}\!,\! B_i^{(t)})$
   \EndFor
   \State \textbf{Return} Domain weights $\Bar{\valpha} = \frac{1}{T}\sum_{t=1}^T \valpha^{(t)}$ 
\end{algorithmic}
\end{algorithm}

\begin{algorithm}[htpb!]
   \caption{\doge Domain Reweighting (for Out-of-domain Generalization).}
   \label{alg:doge-ood}
\begin{algorithmic}[1]
   \State {\bfseries Input:} Training domain data splits $D_1,\dots, D_k$, \textcolor{DarkBlue}{OoD domain $D_{ood}$}, Proxy model weights $\vtheta^{(0)}$, Hyperparameters: number of training steps $T$, batch size $b$, step size $\eta^{(t)}$, Bregman coefficient $\mu$. 
   \State Initialize proxy weights $\vtheta^{(0)}$
   \State Initialize proxy domain weights $\valpha^{(0)} = \frac{1}{k}\mathbf{1}$
   \For{$t \in [T]$}
        \State \emph{Uniformly} draw $B^{(t)}=\{B_1^{(t)},\dots,B_k^{(t)}\} \textcolor{DarkBlue}{\cup \{B_{ood}^{(t)}\}}$ 
        \State Obtain $\nabla l_i(\vtheta^{(t)}, B_i^{(t)})$ for $i \in [k]$ 
        \State \textcolor{DarkBlue}{Obtain $\nabla l_{ood}(\vtheta^{(t)}, B_{ood}^{(t)})$} \hfill 
        \State Compute \textcolor{DarkBlue}{$\gW^{(t)}_{ood}$}
        \State Update domain weights according to Eq.~\eqref{equ:update_dw}:\\
        $\qquad \qquad \displaystyle{\hat{\valpha}^{(t)} \gets \valpha^{(t-1)} \odot \exp(\eta^{(t)}\textcolor{DarkBlue}{\gW^{(t)}_{ood}}/\mu)}$ \\
        $\qquad \qquad \valpha^{(t)} \gets \displaystyle{\hat{\valpha}^{(t)}/\sum_{i=1}^k \hat{\alpha}^{(t)}_i}$
        \State Update $\vtheta^{(t)}$:\\
        $\qquad \ \vtheta^{(t+1)} = \vtheta^{(t)} - \eta^{(t)} \sum_{i\in [k]}\alpha_i^{(t)} \nabla l_i(\vtheta^{(t)}, B_i^{(t)})$
   \EndFor
   \State \textbf{Return} Domain weights $\Bar{\valpha} = \frac{1}{T}\sum_{t=1}^T \valpha^{(t)}$ 
\end{algorithmic}
\end{algorithm}
\textbf{Training the base model.} Given the final domain weights $\Bar{\valpha}$, we train the full size model by sampling $D_{train}$ according to $P_{\Bar{\valpha}} \triangleq  \sum_{i=1}^k \Bar{\alpha}_i \cdot \unif(D_i)$.


\begin{figure}[ht!]
  \centering
  \begin{subfigure}[Domain weights]{
  \includegraphics[width=\linewidth,clip]{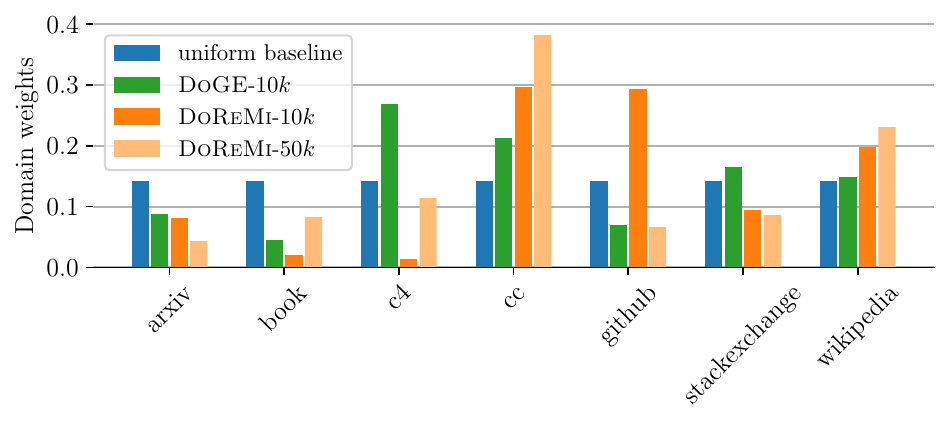} 
  } 
  \end{subfigure} 
  \begin{subfigure}[Average reasoning accuracy]{
  \includegraphics[width=\linewidth,clip]{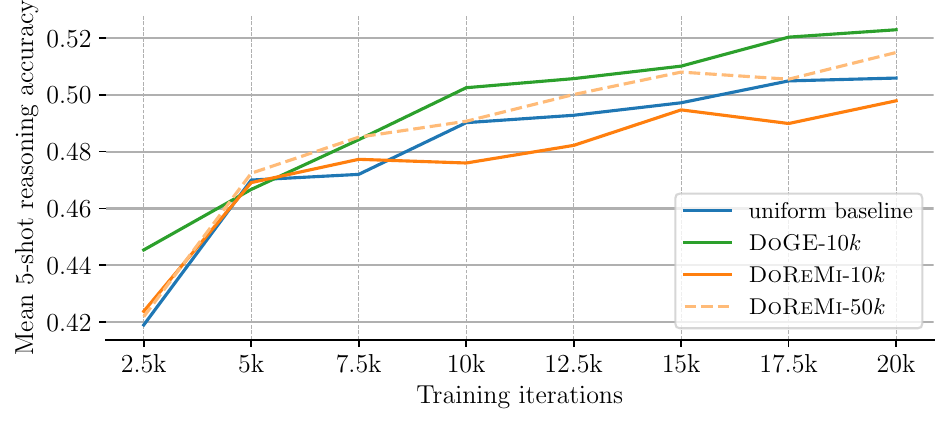}
  }
  \vspace{-1em}
  \end{subfigure} 
 \caption{\textbf{Universal generalization results on SlimPajama.} In \textbf{(a)} we compare several domain weight distributions obtained by \doge and \textsc{DoReMi}. We show two distributions for \textsc{DoReMi} obtained by training the auxiliary models for $10k$ and $50k$ iterations. \doge's proxy model has been trained for $10k$ iterations. In \textbf{(b)} we plot the average 5-shot accuracy during the base model training. \doge acquires few-shot reasoning ability faster than all other baseline methods and improves the final average accuracy by a large margin.}
 \vspace{-1em}
 \label{fig:fewshot}
\end{figure}

\section{\textsc{DOGE} Improves Generalization}\label{sec:experiment}
In this section, we show how \textsc{DoGE} is reweighting the source domains to improves the model's performances in both universal generalization and out-of-domain generalization settings. 

\subsection{Universal Generalization}\label{sec:universal-exp}
In the case of universal generalization, we aim to improve the model's generalization across all domains present in the training set. We measure the average perplexity across all domains and 5-shot reasoning ability across a series of reasoning tasks, covering diverse knowledge fields including physics, social science, logic inference etc.:
COPA~\citep{gordon-etal-2012-copa}, SciQ~\citep{welbl-etal-2017-sciq}, PIQA~\citep{bisk2019piqa}, LogiQA~\citep{liu2020logiqa}, WiC~\citep{pilehvar2019wic} and WinoGrande~\citep{sakaguchi2019winogrande}. We use \emph{LM-eval Harness}~\citep{eval-harness} to assess the few-shot reasoning performance. 

\textbf{Training setup.}  
We experiment on SlimPajama~\citep{cerebras2023slimpajama}, which is a deduplicated version of RedPajama consisting of data from $7$ domains. 
We train a small $82$M decoder-only transformer~\citep{vaswani2023attention} as the \emph{proxy model} for domain reweighting. Auxiliary models for both \doge and \textsc{DoReMi} are trained for $10k$ iterations. 
We also experiment with training the auxiliary models of \textsc{DoReMi} for $50k$ steps, giving that baseline a strong advantage. 
The final domain weights are used to train larger \emph{base models} ($124$M, $210$M, $684$M). We refer to those three methods as $\textsc{DoGE-}10k$, $\textsc{DoReMi-}10k$ and $\textsc{DoReMi-}50k$. We also compare to the \textsc{Baseline} with uniform domain weights, which is the best heuristic for universal generalization without prior knowledge on inter-domain relatedness. We report domain weights from $\textsc{DoGE-}10k$ as the average of three random seeds. All models are trained from scratch with batch size of $128$, and sequence length of $512$. The vocabulary size of the tokenizer is $50304$. Details on model architectures are provided in App.~\ref{apdx:model-archs}. 


\ignore{\begin{figure*}[ht!]
    \centering
    \includegraphics[width=\textwidth]{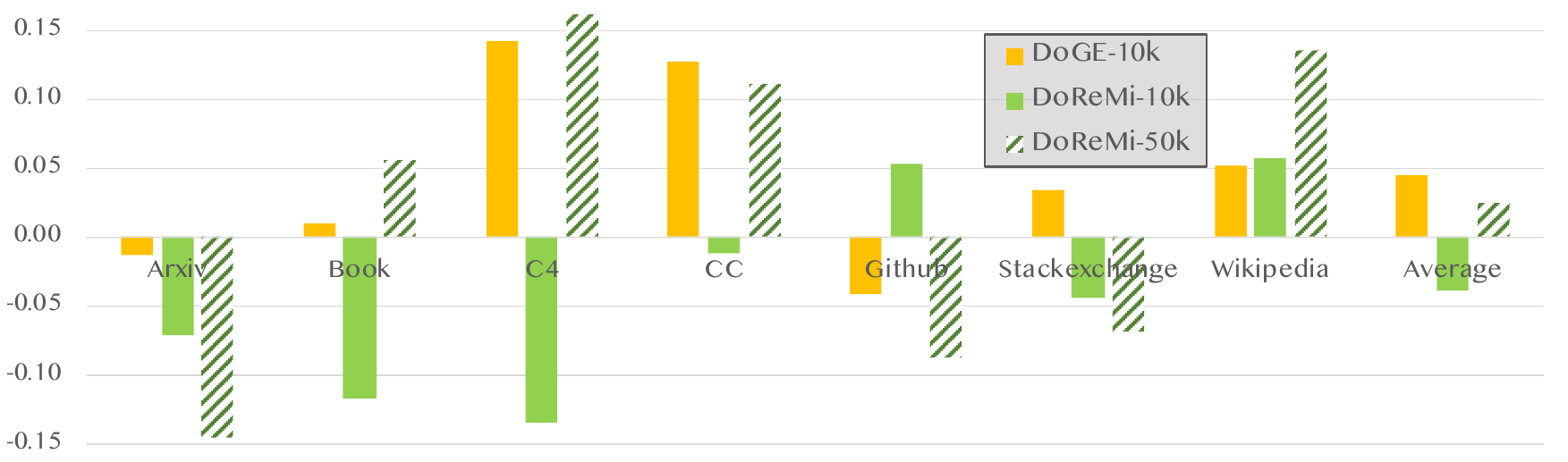}
    \vspace{-1.5em}
    \caption{\textbf{Perdomain Log-Perplexity improvement from \texttt{Baseline} with uniform domain weights.} $y$-axis denotes the reduction of log-perplexity from \texttt{Baseline}, the higher the better. We compare \doge with \textsc{DoReMi}-$10k$, \textsc{DoReMi}-$50k$. \doge outperforms the uniform baseline on 5 out of 7 domains in SlimPajama and achieves the best average perplexity over all baseline methods.} 
    \label{fig:ppl}
\end{figure*} }

\begin{table*}[ht!]
\caption{\textbf{Per-domain Perplexity for universal generalization with $684$M parameter models.} We compare \doge-$10k$ with \textsc{DoReMi}-$10k$, \textsc{DoReMi}-$50k$ and a Baseline with uniform domain weights. We measure the perplexities on validation sets for all of the $7$ domains of SlimPajama. \doge outperforms the uniform baseline on $5$ out of $7$ domains and achieves the best average perplexity over all baseline methods. Scores outperforming the baseline are in \textbf{Bold}. The average perplexity is calculated as the exponential of the average loss across all domains instead of the average of all domain perplexities.}
\vspace{0.5em}
\label{tab:universal_ppl}
\centering
\small 
\begin{adjustbox}{max width=0.95\textwidth}
\begin{tabular}{lccc||c}
\toprule
                    Domain & Uniform baseline & \textsc{DoReMi}-$10k$ & \textsc{DoGE}-$10k$ & \textsc{DoReMi}-$50k$ \\
                    \midrule
Arxiv     &  \textbf{8.105}  & 8.698 & 8.207 & 9.378 \\
Book   &   44.990  & 50.594 & \textbf{44.574} & 42.557 \\
C4 &  49.066 & 56.116 & \textbf{42.558} & 41.388 \\
CommonCrawl & 45.903 & 46.459 & \textbf{40.432} & 41.067 \\
Github &  3.944 & \textbf{3.739} & 4.107 & 4.301 \\
Stackexchange &  8.628 & 9.022 & \textbf{8.332} & 9.235 \\
Wikipedia &  12.047 & \textbf{11.380} & 11.443 & 10.519 \\
\midrule
Average &  16.526 & 17.172 & \textbf{15.806} & 16.124 \\
Worst-case &  49.066 & 56.116 & \textbf{44.574} & 42.557 \\
\midrule
\# domains outperform Baseline &  / & 2 & 5 & 4\\
\bottomrule
\end{tabular}
\end{adjustbox}
\end{table*}

\ignore{\begin{table*}[htp!]
\caption{\textbf{Perdomain Perplexity for universal generalization.} We compare \doge with \textsc{DoReMi}-$10k$, \textsc{DoReMi}-$50k$ and Baseline with uniform domain weights on the average validation perplexity across all 7 domains in SlimPajama. \doge outperforms the uniform baseline on 5 out of 7 domains and achieves the best average perplexity over all baseline methods. Performance better than \texttt{Baseline} are in \textsc{Bold}.}
\vspace{0.2em}
\label{tab:universal_ppl}
\centering
\begin{adjustbox}{max width=0.95\textwidth}
\begin{tabular}{lcccc}
\toprule
                    Domain & Baseline & \textsc{DoReMi}-$10k$ & \textsc{DoReMi}-$50k$ & \textsc{DoGE}\\
                    \midrule
Arxiv     &  \textbf{8.105}  & 8.698 & 9.378 & 8.207\\
Book   &   44.990  & 50.594 & \textbf{42.557} & \textbf{44.574}\\
C4 &  49.066 & 56.116 & \textbf{41.388} & \textbf{42.558}\\
CommonCrawl & 45.903 & 46.459 & \textbf{41.067} & \textbf{40.432}\\
Github &  3.944 & \textbf{3.739} & 4.301 & 4.107\\
Stackexchange &  8.628 & 9.022 & 9.235 & \textbf{8.332}\\
Wikipedia &  12.047 & \textbf{11.380} & 10.519 & 11.443\\
\midrule
Average\footnote{The average perplexity is calculated as the exponential of the average loss across all domains instead of the average of all domain perplexity.} &  16.526 & 17.172 & 16.124 & \textbf{15.806}\\
\midrule
\# domains outperform Baseline &  / & 2 & 4 & 5\\
\bottomrule
\end{tabular}
\end{adjustbox}
\end{table*}}

\begin{table*}[htbp!]
\caption{\textbf{Exact-match accuracies(\%) for $5$-shot reasoning tasks.} In $5$ out of $6$ tasks, \doge reaches the best accuracy compared to other baseline methods. Only \textsc{DoReMi}-$50k$ slightly outperforms \doge on PIQA, using $40k$ more steps to train the auxiliary models.  
}
\vspace{0.5em}
\label{tab:fewshot}
\small
\centering
\begin{adjustbox}{max width=\textwidth}
\begin{tabular}{lccc||c}
\toprule
                 Task & Uniform baseline & \textsc{DoReMi}-$10k$ & \textsc{DoGE}-$10k$ & \textsc{DoReMi}-$50k$ \\
                 \midrule
COPA   &        58.00    &  59.00 & \textbf{62.00} &  61.00         \\
SciQ   &        61.80    &  60.30 & \textbf{65.00} &  64.50        \\
LogiQA &        23.20    &  24.58 & \textbf{25.50} &  23.81        \\
PIQA   &        59.85    &  56.86 & 60.34 &  \textbf{60.94}        \\
WiC    &        \textbf{49.69}    &  48.59 & \textbf{49.69} &  49.53        \\
WinoGrande&     50.99     &  49.41 & \textbf{51.22} &  49.17         \\
\midrule
Average   &       50.59         &    49.79      & \textbf{52.29} &  51.49                 \\
\bottomrule
\end{tabular}
\end{adjustbox}
\end{table*}

\textbf{Evaluation on language modeling ability.} We measure the per-domain perplexity on held-out validation sets for the largest scale base model ($684$M). Results for other model sizes ($124$M and $210$M) are provided in App.~\ref{apdx:universal-eval}. 
According to Tab.~\ref{tab:universal_ppl}, \doge-$10k$ outperforms \textsc{Baseline} and \textsc{DoReMi}-$10k$ in 5 out of 7 domains, by a large margin. Notably, \doge-$10k$ outperforms all the other baseline methods in terms of average perplexity, given a great advantage in the number of iterations to train \textsc{DoReMi}-$50k$.

\textbf{Evaluation on few-shot reasoning accuracy.} We test the $5$-shot reasoning accuracy across $6$ tasks for our largest ($684$M) models. According to Tab.~\ref{tab:fewshot} and Fig.~\ref{fig:fewshot}.(b), \doge-$10k$ improves few-shot reasoning ability of the \emph{base model}, especially at the early training stage. Our method outperforms the \textsc{Baseline} and \textsc{DoReMi}-$10k$ on all $6$ tasks. 
In contrast, \textsc{DoReMi}-$10k$ slightly hurts the reasoning accuracy. With $40k$ more training iterations, \textsc{DoReMi}-$50k$ outperforms uniform \textsc{Baseline} on most of the tasks, while still left behind \doge-$10k$ on $5$ out of $6$ reasoning tasks. 
On average, \doge-$10k$ improves the $5$-shot reasoning ability by $1.7$ accuracy points over uniform \textsc{Baseline}, which outperforms all the other methods. 


\begin{table*}[htbp!]
\vspace{-0.1in}
\caption{\textbf{Out-of-Domain generalization results on SlimPajama.} \doge significantly outperforms the uniform averaging baseline in all domains. This demonstrates \doge's ability to select helpful source data for the given target. Domain weights for each target domain are present in Fig.~\ref{fig:dw_ood}.c. Even when finetuning the pretrained models on the target domain, \doge models reach a better perplexity. Performance better than the baseline are highlighted in \textbf{bold}.}
\vspace{0.2em}
\label{tab:ppl-ood}
\centering
\begin{adjustbox}{max width=0.95\textwidth}
\begin{tabular}{lcc||cc||c}
\toprule
                  & Baseline (w$/$o target) & \textsc{DoGE} & Baseline (w$\slash$o target)\small+fine-tuning & \textsc{DoGE}\small+fine-tuning & Oracle (with target)\\ 
                  \midrule
Arxiv             & 18.92\small±0.14    &     \textbf{16.70}\small±0.08    & 10.47\small±0.01 & \textbf{10.20}\small±0.01 & 9.78\small±0.01\\
Book              & 82.57\small±0.05    &     \textbf{63.89}\small±0.18    & 65.73\small±0.06 & \textbf{56.94}\small±0.24& 66.43\small±0.19    \\
C4                & 89.56\small±0.38    &     \textbf{63.96}\small±0.11    & 71.24\small±0.09 & \textbf{56.91}\small±0.17& 70.69\small±0.14    \\
CommonCrawl       & 81.65\small±0.47    &     \textbf{57.77}\small±0.56    & 65.75\small±0.01 & \textbf{51.173}\small±0.04& 67.06\small±0.15    \\
Github            & 6.675\small±0.00     &     \textbf{5.091}\small±0.03     & 4.99\small±0.01 & \textbf{4.26}\small±0.01& 4.97\small±0.01     \\
StackExchange     & 16.941\small±0.02    &     \textbf{14.77}\small±0.01    & 11.24\small±0.004 & \textbf{10.98}\small±0.002 & 11.26\small±0.03    \\
Wikipedia         & 58.04\small±0.32    &     \textbf{53.87}\small±0.35    & 18.38\small±0.02 & \textbf{17.71}\small±0.05 & 17.61\small±0.02    \\
\bottomrule
\end{tabular}
\end{adjustbox}
\end{table*}

\begin{figure*}[ht!]
    \centering
    \includegraphics[width=\textwidth]{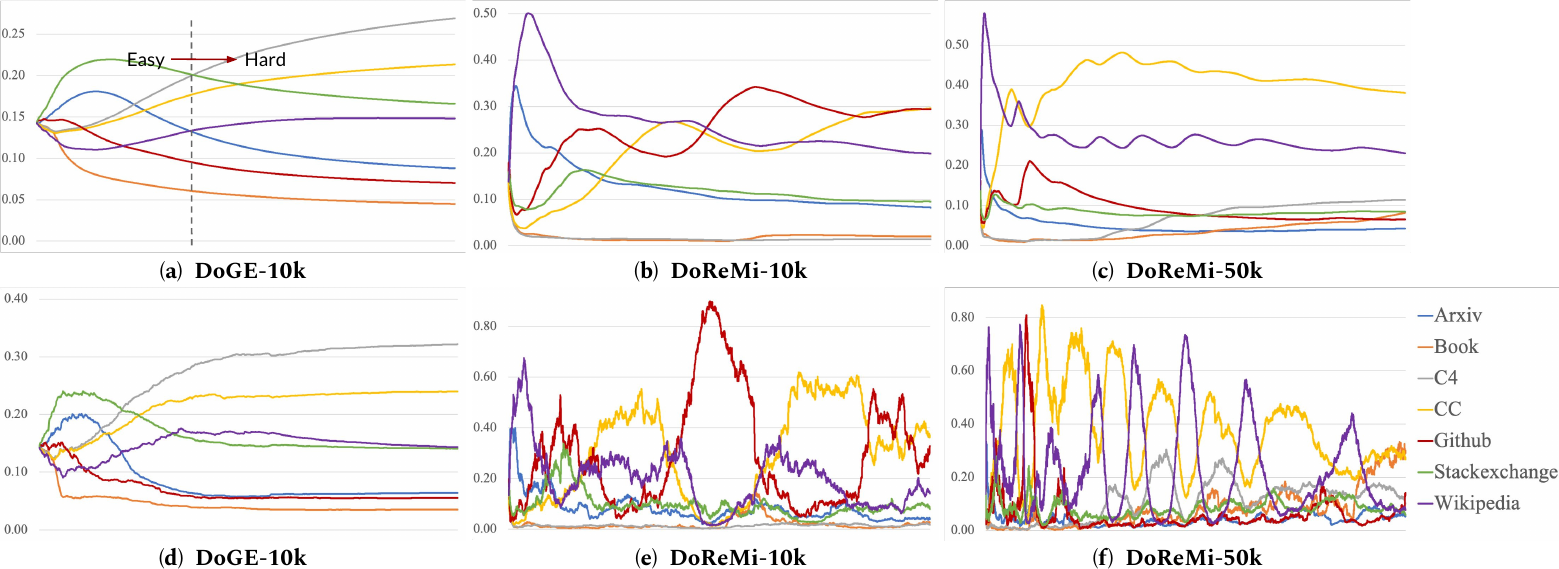}
    \vspace{-1em}
    \caption{\textbf{Average (Top row) and step-wise (Bottom row) domain weights evolution.} We train the auxiliary models of \textsc{DoReMi} for $10k$ or $50k$ steps, which yields \textsc{DoReMi}-$10k$ and \textsc{DoReMi}-$50k$. In \textbf{\textsc{DoGE} (a, d)}, we observe a clear two-phase curriculum from \emph{easy} to \emph{hard} domains. In contrast, the step-wise domain weights from both \textbf{\textsc{DoReMi}-$10k$ (e)} and \textbf{\textsc{DoReMi}-$50k$ (f)} oscillate greatly during training. The final average domain weights differ a lot between \textbf{\textsc{DoReMi}-$10k$ (b)} and \textbf{\textsc{DoReMi}-$50k$ (c)}, which reveals a strong dependency on the capacity of auxiliary models and the training iterations.} 
    \label{fig:dw}
\end{figure*}

\textbf{Evolution of domain weights.}
Fig.~\ref{fig:dw} shows the step-wise \textbf{(Bottom)} and average \textbf{(Top)} domain weights evolution during the training of the proxy model. The step-wise domain weights can be interpreted as the online contributions from each domains, while the final domain weights $\Bar{\valpha}$ are given by the average. According to Fig.~\ref{fig:dw}.a and Fig.~\ref{fig:dw}.d, \doge shows a clear phase transition, with different stages of training, as in a curriculum: in an early stage, \doge up-weights Arxiv and Stackexchange while gradually up-weighting C4, CC and Wikipedia, which contain a more diverse lexical coverage and complicated semantics. 
The domain-weights for the other two domains (GitHub and Book) are kept low. We hypothesise that Github---with its emphasis on code---has limited vocabulary and semantic knowledge, and the complexity of Book might be covered by C4 and CC, which are the two most up-weighted domains.
In comparison, the step-wise dynamic of \textsc{DoReMi}-$10k$ in Fig.~\ref{fig:dw}.e and \textsc{DoReMi}-$50k$ in Fig.~\ref{fig:dw}.f oscillate greatly during training. Despite the additional training steps for both auxiliary models in \textsc{DoReMi}-$50k$, the final average domain weights differ greatly with the ones from \textsc{DoReMi}-$10k$ (by a mean absolute difference of $0.08$), which indicates the strong dependency of \textsc{DoReMi} on the capacity of the reference model and training iterations. We present the final domain weights adopted by different methods in Fig.~\ref{fig:fewshot}.a.

\textbf{Robustness to the scale of proxy model.}
To further explore the how the scale of the proxy model could impact the final domain weights, we run the ablation experiments on three different scales ($60$M, $82$M, $124$M). Notably, \doge's final domain weights are consistent across various scale of proxy model. The mean absolute difference of domain weights between $60$M (resp. $124$M) and $82$M proxy models is less than $0.015$ (resp. $0.005$) across all $7$ domains, which demonstrates the robustness of our method. Compared to \textsc{DoReMi}, \doge has less dependencies on the capacity on the auxiliary model(s) which requires less efforts and costs to tune the size of the proxy model and choose the number of iterations. The details of the ablation experiments are presented in App.~\ref{apdx:robustness}.

\textbf{Comparison of Computation Overhead.} Our experiments show \doge to be more memory, time, and data efficient than \textsc{DoReMi}. Indeed, \textsc{DoReMi} requires two auxiliary models of the same scale, while \doge only requires a single proxy model. Moreover, while $10k$ steps were sufficient for \doge to improve the perplexity and few-shot reasoning accuracies over the uniform baseline, \textsc{DoReMi} required $5\times$ more tokens and $10\times$ more floating point operations. 

\subsection{Out-of-Domain Generalization}\label{sec:ood-exp}

In the case of Out-of-Domain (OoD) generalization, we aim to improve the model's generalization to a target domain which is not part of the training mixture $D_{train}$. 
Given the target domain is missing from $D_{train}$, we expect \doge to up-weigh the helpful domains among $D_{train}$ while sampling less from distinct ones. We consider two dataset: SlimPajama and Wiki40b. Since \textsc{DoReMi} does not support this use-case, we only compare \doge with \textsc{Baseline} with uniform domain weights. The \textsc{Oracle} baseline also enables access to the target domain, which shares the same sampling weight as other source domains in $D_{train}$. We assess the target domain perplexity on the held-out test set and report average results over two seeds.

\textbf{Wiki40b setup.} We test the OOD-generalization capabilities of \doge in a multilingual setting, aiming to facilitate low-resource language learning from mainstream language corpus. We use the Wiki40b dataset \citep{guo-etal-2020-wiki}, which consists in a collection of Wikipedia articles in $40+$ languages. We set English, German, Spanish, French and Russian as source domains in $D_{train}$. The target domain is set to Catalan or Dutch, which are considered as low-resource languages. We train the proxy model ($82$M) for $10k$ steps to obtain the domain weights and then train the base model ($124$M) for $10k$ steps.

\textbf{SlimPajama-OoD setup.} For out-of-domain generalization, we set each of the domains in SlimPajama as the target domain, with $0.05$B tokens accessible. The remaining 6 domains are used as source domains $D_{train}$, each with $2$B tokens accessible. We run the proxy model ($82$M) for $10k$ steps to obtain the domain weights and then train the base model ($124$M) for $10k$ steps. We continually fine-tunine the pretrained checkpoints for $1000$ steps on the target domain.


\textbf{Perplexity on the target domain.} 
In Tab.~\ref{tab:ppl-ood}, we show how \doge consistently outperforms the uniform baseline across all seven domains in SlimPajama. On C4 and CommonCrawl, \doge achieves a better performance than the oracle without further fine-tuning. This demonstrates that irrelevant data sources can deteriorate the adaptation to the target domain, and \doge can help to select helpful source domains. 
After finetuning the pretrained checkpoints on the target domain, \doge pretrained models still outperform the finetuned baseline. 
In Fig.~\ref{fig:dw_ood}.a and Fig.~\ref{fig:dw_ood}.b, we show the test perplexity on Catalan and Dutch when training the base models. \doge models show a significant improvement over the uniform baseline by learning from related mainstream languages. 

\ignore{
\begin{figure}[ht!]
  \centering
  \begin{subfigure}[Universal Generalization]{
  \includegraphics[width=\linewidth,clip]{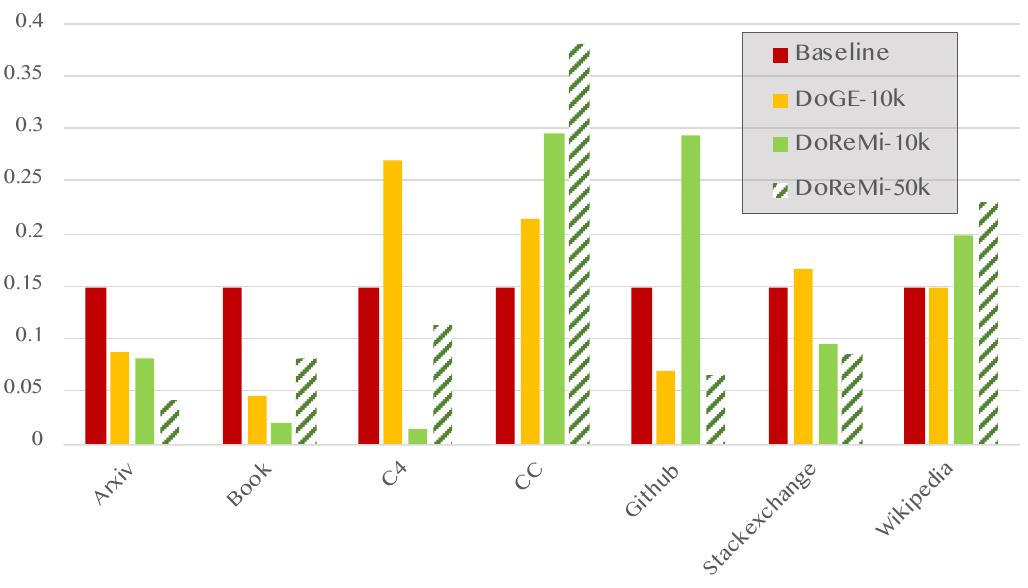} 
  } 
  \end{subfigure} 
  \begin{subfigure}[Out-of-domain Generalization]{
  \includegraphics[width=\linewidth,clip]{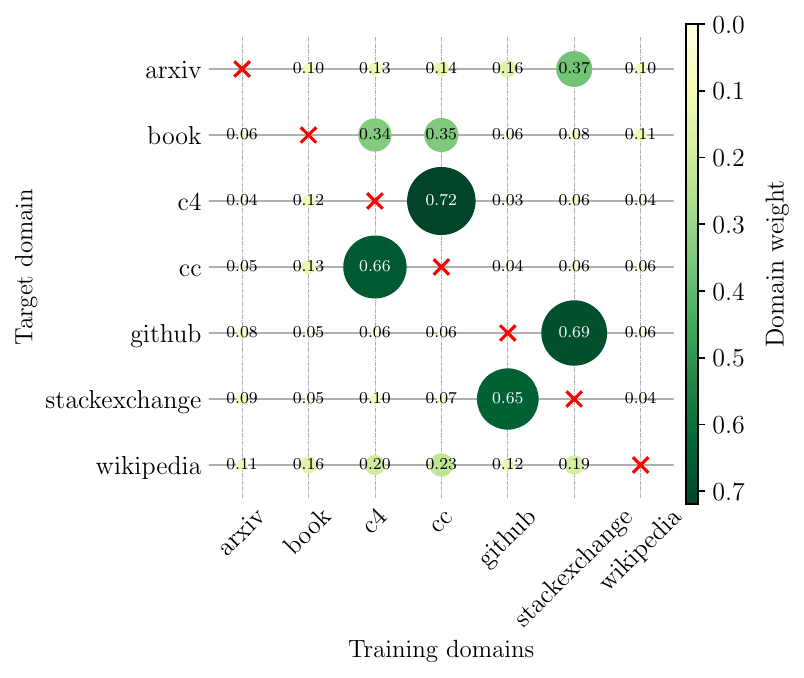} 
  } 
  \end{subfigure} 
 \caption{\textbf{Domain weights for universal and out-of-domain generalization.} }
 \label{fig:dw_ood}
\end{figure}
} 

\begin{figure}[ht!]
  \centering
  \begin{minipage}{0.6\textwidth}
      \begin{subfigure}[OoD generalization to Catalan]{
      \begin{adjustbox}{valign=t}
      \includegraphics[width=0.53\linewidth,clip]{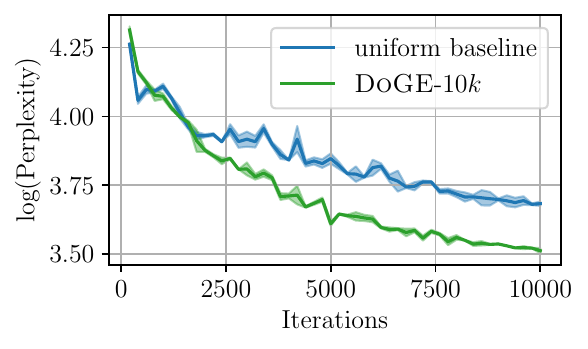} 
      \end{adjustbox}
      \begin{adjustbox}{valign=t}
      \includegraphics[width=0.22\linewidth,clip]{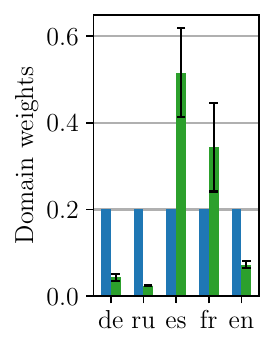} 
      \end{adjustbox}
      } 
      \end{subfigure} 
  \end{minipage}
  \begin{minipage}{0.6\textwidth}
      \begin{subfigure}[OoD generalization to Dutch]{
      \begin{adjustbox}{valign=t}
      \includegraphics[width=0.53\linewidth,clip]{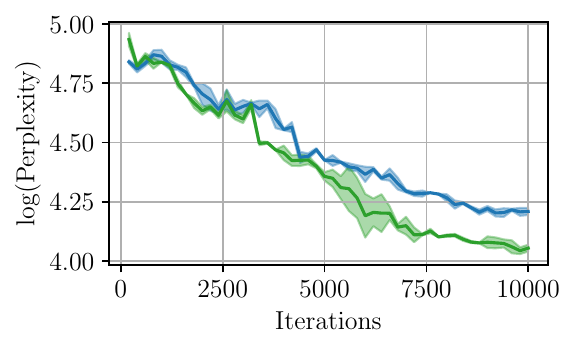} 
      \end{adjustbox}
      \begin{adjustbox}{valign=t}
      \includegraphics[width=0.22\linewidth,clip]{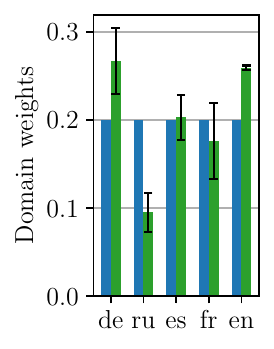} 
      \end{adjustbox}
      } 
      \end{subfigure} 
      \vspace{-1em}
  \end{minipage}
  \begin{subfigure}[OoD generalization weights on SlimPajama]{
  \includegraphics[width=0.98\linewidth,clip]{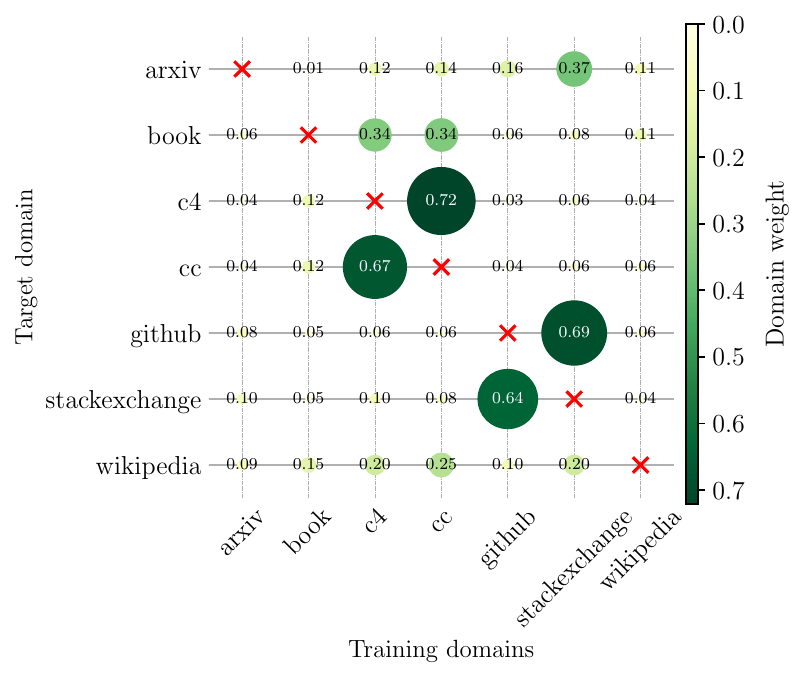} 
  } 
  \end{subfigure} 
  \vspace{-1em}
 \caption{\textbf{Out-of-Domain generalization results.} In \textbf{(a)} (resp. \textbf{(b)}), we compare \doge with the uniform domain weights baseline when attempting to generalize to Catalan (resp. Dutch) from a mixture of German, Russian, Spanish, French and English wikipedia articles from the Wiki40b dataset. The target languages are absent from the training mixture. The histograms show \doge up-weights training languages having some similarity with the target. As a result, \doge's loss on the target is decreasing faster than the baseline. In \textbf{(c)} we show \doge domain weights for OoD generalization on the SlimPajama dataset. Those domain weights result from training mixtures consisting of all the training domains except for one which is used as the target ($D_{ood}$) domain. Each row represents a distribution returned by Alg.~\ref{alg:doge-ood}. The target domain is not used during training and hence is marked by a red cross. The weight distributions look very coherent, e.g. to generalize to GitHub, \doge upweights stackexchange which contains a significant fraction of code. Similarly, to generalize to cc, the c4 domain---which also consists in web data---is up-weighted. }
 \label{fig:dw_ood}
\end{figure}

\textbf{Automatically detected inter-domain affinities.}
For the SlimPajama experiments, we presents the auto-detected inter-domain affinities obtained by \doge in Fig.~\ref{fig:dw_ood}.c. There is a clear inter-dependency between C4 and CommonCrawl, which are both web-crawled data sources; meanwhile, the strong affinity between Stackexchange and Github is also detected, which both contain code-related knowledge. 
Similarly, the domain weights obtained in multilingual experiments reflects the languages relatedness in etymology~\citep{langauge-tree}, where Catalan is close to French and Spanish in Italic family, while Dutch is close to German in Germanic family. (Fig.~\ref{fig:dw_ood}.a and Fig.~\ref{fig:dw_ood}.b). 


\section{Discussion and Limitations}\label{sec:discussion}
\textbf{Stage-wise domain weights is no better than global average.}
Following the success of curriculum learning \cite{hacohen2019power,xu-etal-2020-curriculum,fan2023irreducible} in multiple fields, we explore the potential of applying stage-wise time-varying domain weights during the training of the base model. We manually divided the training process of the proxy model into $K=2,3,10$ stages and average the step-wise domain weights respectively to be the stage-wise domain weights (Fig. \ref{fig:curriculum_dw}). 
By applying stage-wise domain weights, the total amount of samples from each domain are the same as the global domain weights.
As shown in App.~\ref{apdx:curriculum}, none of the time-varying strategies show clear improvement over the global averaged domain weights in average perplexity. However, with $K=2,3$, the stage-wise domain weights help the model learn \emph{hard} domains (Wikipedia, CC, C4) better, which aligns with the principle of curriculum learning. With $K=10$, the domain weights updates every $1000$ steps, while the performance of the base model is much worse than applying static domain weights.

\textbf{The proxy model performs worse than a same-scale base model.} 
With the step-wise dynamic adaptation, it is expected that the proxy model with the rescaled gradient could outperform the base model trained with the learned domain weights. However, compared with a base model with the same scale ($82$M), we find that the proxy model consistently performs worse in validation perplexity (see App.~\ref{apdx:proxy}). A similar behavior is also mentioned by \citet{xie2023doremi}, where both auxiliary models (the reference and the proxy) in \textsc{DoReMi} cannot reach  comparable performance to the same-scale base model with resampling. 


\textbf{Better efficiency using parameter selection.}
The computation budget for generalization estimation $\gW$ is quadratic to the scale of model. Thus, we explore the potential of parameter selection based on cancellation effect following the empirical success of \citep{yeh2022better}. Specifically, we rank all parameter modules (i.e. transformer blocks or embedding layers) of model weights by the cancellation effect and only use gradients of the selected modules when compute $\gW$. Among the five parameter selection strategies, selecting $30$ modules with highest cancellation effect achieve the comparable average perplexity with only $2.5\%$ computation costs for generalization estimation $\gW$. We provide the details of parameter selection in Appendix (\S~\ref{apdx:cancellation}).


\section{Related Work}\label{sec:related-work}
\textbf{Data Selection for Language Modeling.} Many works show how a rigorously selected training corpus can effectively improve downstream performance with fewer training tokens. 
\citet{longpre2023pretrainers} discover a trade-off between a model's toxic generalization behavior and its generalization ability by applying quality control with various thresholds. 
\citet{gunasekar2023textbooks} and \citet{li2023textbooks} trained a $1.3$B model \textsc{Phi-1} using $7$B \emph{text-book quality} code data, outperforming previous larger models trained on larger dataset, illustrating the potential of high-quality data. 

However, due to scalability issues, most traditional data selection methods fail to be applicable for pretraining. Classifier-based data filtering techniques are commonly used to construct a pretraining corpus \citep{gao2020pile,penedo2023refinedweb}. 
\citet{everaert2023gio}  propose GIO to select a subset that minimizes the KL-divergence to the target distribution, yet incurs high computation complexity. \citet{xie2023dsir} present a scalable importance resampling strategy by reducing dimensionality into an n-gram-featured subspace, which risks from a weak representation for sophisticated semantics. \citet{engstrom2024dsdm} train a linear datamodel first to predict a mapping from training dataset to downstream loss, then select a subset to minimize the approximated loss. 

\textbf{Data Reweighting for LLM Pretraining.}
Instead of selecting a subset, data reweighting remain the full access to the whole dataset while re-scale the contribution of each instance under various target tasks. 
\citet{grangier2023adaptive} train an extra weighting network to re-weight the loss from each data point using bilevel optimization algorithms. \citet{thakkar2023selfinfluence} measure self-influence as the sample importance during pretraining.
Compared to instance-wise strategies, domain reweighting aims to reweigh or resample from various data groups, which offers better scalability for language model pretraining. 
\textsc{DoReMi}~\citep{xie2023doremi} applies Group DRO on the loss gap between two auxiliary models to optimize the domain sampling weights. 
\citet{chen2023skillit} propose to build an online resampling curriculum by exploiting the dependency relationship among skills represented by a directed skill graph. While the computation cost for constructing the skill graph limits its applicability to general language model pretraining.  

\section{Conclusion}\label{sec:conclusion}
We introduced \doge, an effective and efficient domain reweighting framework based on generalization estimation, which finds the optimal domain weights tailored to various generalization objectives. 
With the pretraining corpus with reweighted domain sampling weights, our experiments on SlimPajama show an improvment on LLM's universal generalization on langauge modelling and downstream few-shot reasoning ability. 
With out-of-domain generalization objective, \doge efficiently accelerates the learning of the target domains and low-resource language by selectively learning from related data sources. Notably, \doge gives robust domain reweighting results across various scales of proxy models, which demonstrates a great capacity to utilize small-scale proxy model to accelerate the training of larger models.
Scaling-up experiments with larger models and datasets is an important future direction.

\newpage
\section{Impact Statement} 
This paper presents work whose goal is to advance the field of Machine Learning. There are many potential societal consequences of our work, none which we feel must be specifically highlighted here.
\bibliography{example_paper}
\bibliographystyle{icml2023}

\newpage
\appendix
\onecolumn
\section{Model Architectures}\label{apdx:model-archs}
The  maximal (min.) learning rate applied to train the largest model ($684$M) is $1.5\times 10^{-4}$ ($5\times 10^{-5}$), while others apply $5\times 10^{-4}$ ($1\times 10^{-4}$), with a cosine scheduler. The weight decay for all models is set as 0.01, the gradient clip is set as 1.0. 
\begin{table}[htbp!]
\caption{Architecture hyperparameters for various model scales used in the paper. All models are vanilla Transformer decoder-only models.}
\label{tab:archictectures}
\centering
\vspace{5pt}
\begin{adjustbox}{max width=0.9\textwidth}
\begin{tabular}{lccccc}
\toprule
     & Layers & Attention heads & Embed dim & Hidden dim & Max. learning rate (min.) \\
     \midrule
60M & 3      & 6               & 768       & 3072    & $5\times 10^{-4}$ ($1\times 10^{-4}$)   \\     
82M & 6      & 12               & 768       & 3072    & $5\times 10^{-4}$ ($1\times 10^{-4}$)   \\
124M & 12     & 12              & 768       & 3072    & $5\times 10^{-4}$ ($1\times 10^{-4}$)  \\
210M & 24     & 16              & 768       & 3072    & $5\times 10^{-4}$ ($1\times 10^{-4}$)  \\
684M & 36     & 24              & 1200       & 4800   & $1.5\times 10^{-4}$ ($5\times 10^{-5}$)   \\ \bottomrule
\end{tabular}
\end{adjustbox}
\end{table}

\section{Derivation of Domain Weights Update Rule}\label{apdx:algorithm}
To realize the optimal \emph{universal generalization} performance within $\mathcal{T}$ steps, we optimize $\alpha_t$ at each training step $t$, which minimizes averaged cross-entropy loss $\Bar{L}(\vtheta_{T})$ across all $k$ domains at the final stage. 
Denote $l(\vtheta)$ as the next-token prediction (cross-entropy) loss of model parameterized by $\vtheta$, $l_i(\vtheta)$ as the loss of the $i^{th}$ domain $D_i$, our final objective can be written as:
\begin{align}\label{apd-equ:obj-T}
    \min_{\alpha_1, \hdots, \alpha_T \in \Delta^{k}} \Bar{L}(\vtheta^{(T)}) = \min_{\alpha_1, \hdots, \alpha_T \in \Delta^{k}} \sum_{i \in [k]} l_i(\vtheta^{(T)})
\end{align}
With a greedy approximation of~\eqref{apd-equ:obj-T}, we search for the optimal domain weights $\alpha_t$ to minimize the average loss over $k$ domains at step (t+1):
\begin{align}\label{apd-equ:obj-t0}
    \argmin_{\alpha_t \in \Delta^{k}} \Bar{l}(\vtheta^{(t+1)}) &= \argmin_{\alpha_t\in \Delta^{k}} \sum_{i \in [k]} l_i(\vtheta^{(t+1)}) \notag\\
&= \argmin_{\alpha_t\in \Delta^{k}} \sum_{i \in [k]} [l_i(\vtheta^{(t+1)}) - l_i(\vtheta^{(t+1)})]
\end{align}
Take the first-order approximation, we estimate the loss for $i^{th}$ domain as: 
\begin{align}
    l_i(\vtheta^{(t+1)}) &= l_i(\vtheta^{(t)}) + \nabla l_i(\vtheta^{(t)}) \cdot (\vtheta^{(t+1)} -\vtheta^{(t)}) + o(\|\vtheta^{(t+1)} - \vtheta^{(t)}\|) \notag \\
&= l_i(\vtheta^{(t)}) + \nabla l_i(\vtheta^{(t)}) \cdot \left[ -\eta_t \cdot \sum_{j\in [k]} \alpha_T^j \nabla l_j(\vtheta^{(t)})\right]+\vepsilon^{(t)}, \notag
\end{align}
where $\vepsilon^{(t)}=o(\|\vtheta^{(t+1)} - \vtheta^{(t)}\|)=o(\|\valpha\|)$ as the high-order remainder from Taylor expansion. Denote $\Grad_t^i := \nabla l_i(\vtheta^{(t)})$, $W^{(t)}_j \triangleq \langle\nabla l_j(\vtheta^{(t)}), \sum_{i \in [k]}\nabla l_i(\vtheta^{(t)}) \rangle$. 
We write $\gW^{(t)}=[W^{(t)}_1, \ldots, W^{(t)}_k] \in\R^k$ for the score vector regrouping generalization estimations across all domains. Equ. \eqref{apd-equ:obj-t0} can be written as:
\begin{align}\label{apd-equ:obj-t1}
    \min_{\alpha_t \in \Delta^{k}} \Bar{l}(\vtheta^{(t+1)}) &= \min_{\alpha_t\in \Delta^{k}} \eta_t \cdot\sum_{i \in [k]} l_i(\vtheta^{(t+1)}) \notag\\
&\approx \min_{\alpha_t\in \Delta^{k}} -\eta_t \cdot\sum_{i \in [k]}\Grad_t^i \left( \sum_{j \in [k]} \alpha_t^j\Grad_t^j \right) \notag\\
&= \min_{\alpha_t\in \Delta^{k}} -\eta_t \cdot\sum_{i \in [k]} \alpha_t^i \left(\Grad_t^i \sum_{j \in [k]}\Grad_t^j \right) \notag\\
&= \min_{\alpha_t\in \Delta^{k}} -\eta_t \cdot\sum_{i \in [k]} \alpha_t^i W^{(t)}_i \notag\\
&= \min_{\alpha_t\in \Delta^{k}} -\eta_t \cdot \langle \alpha_t, \gW^{(t)} \rangle
\end{align}
We estimate $\vepsilon^{(t)}$ by introducing a regularization term via Bregman divergence $D_h(\alpha || \alpha_{t-1}) = h(\alpha) - h(\alpha_{t-1}) - \langle \nabla h(\alpha_{t-1}), \alpha - \alpha_{t-1} \rangle$, $h(\alpha) = \sum_i \alpha_i \ln \alpha_i$. Adding this to \eqref{apd-equ:obj-t1}, our optimization problem is:
\begin{align}\label{apd-equ:obj-t}
    \alpha_t &:= \argmin_{\alpha \in \Delta^{k}} \Bar{l}(\vtheta^{(t+1)}) \notag\\
        &\approx \argmin_{\alpha \in \Delta^{k}} - \eta_t \cdot \langle \alpha, \Wt \rangle + \mu\cdot D_h(\alpha || \alpha_{t-1}) \notag\\
        &= \argmin_{\alpha \in \Delta^{k}} - \eta_t \cdot\langle \alpha, \Wt \rangle + \mu (h(\alpha) - \langle \nabla h(\alpha_{t-1}), \alpha\rangle)
\end{align}
With $\nabla h(\alpha) =  [\ln \alpha^i + 1]_i$, we take derivative of \eqref{apd-equ:obj-t}:
\begin{align}\label{apd-equ:derivative-t}
    \nabla(\cdot) &= \nabla\left(- \eta_t \cdot \langle \alpha, \Wt \rangle + \mu (h(\alpha) - \langle \nabla h(\alpha_{t-1}), \alpha\rangle)\right) \notag\\
                  &= -\eta_t \cdot\Wt + \mu \cdot [\ln \alpha + 1]_i - \mu \cdot [\ln \alpha_{t-1} + 1]_i = 0 
\end{align}
\begin{equation}
    \Rightarrow \ln \alpha_t := \ln \alpha_* = \ln \alpha_{t-1} + \displaystyle{\frac{\eta_t \Wt}{\mu}}
\end{equation}

\paragraph{Out-of-domain Generalization. ($D_{ood}\notin D_{train}$)}
Here we derive the update rule with the objective to generalize to a target domain, which is not included in the pretraining data sources, i.e. $D_{ood}\notin D_{train}$:
\begin{align}
    \min_{\alpha_1, \hdots, \alpha_T \in \Delta^{k}} \Bar{L}(\vtheta^{(T)}) = \min_{\alpha_1, \hdots, \alpha_T \in \Delta^{k}} l_{ood}(\vtheta^{(T)}) \notag
\end{align}
Denote $\Grad_t^i := \nabla l_i(\vtheta^{(t)})$, we search for the optimal domain weights $\alpha_t$ to minimize the average loss on $D_{ood}$ at step (t+1):
\begin{align}
    \argmin_{\alpha_t \in \Delta^{k}} l_{ood}(\vtheta^{(t+1)}) &= \argmin_{\alpha_t\in \Delta^{k}} l_{ood}(\vtheta^{(t+1)}) \notag\\
&= \argmin_{\alpha_t\in \Delta^{k}} [l_{ood}(\vtheta^{(t+1)}) - l_{ood}(\vtheta^{(t)})] \notag\\
&= \argmin_{\alpha_t\in \Delta^{k}} -\eta_t \cdot\sum_{i \in [k]} \alpha_t^i \left(\Grad_t^i\Grad_t^{ood} \right) +\vepsilon^{(t)}
\end{align}
Alternatively, we define the generalization gain of $i^{th}$ domain for the targeted domain $D_{ood}$ as $\gW^{(t)}_{ood} := \alpha_t^i\nabla l_i(\vtheta^{(t)}) \cdot \nabla l_{ood}(\vtheta^{(t)})$. Therefore, the optimization problem can be written as:
\begin{align}
    \argmin_{\alpha_t \in \Delta^{k}} l_{ood}(\vtheta^{(t+1)}) &= \argmin_{\alpha_t\in \Delta^{k}} -\eta_t \cdot\sum_{i \in [k]} \alpha_t^i \left(\Grad_t^i \Grad_t^{ood} \right)  \notag\\
&= \argmin_{\alpha_t\in \Delta^{k}} -\eta_t \cdot\sum_{i \in [k]} \alpha_t^i \gW^{(t)}_{ood} \notag\\
&= \min_{\alpha_t\in \Delta^{k}} -\eta_t \cdot \langle \alpha_t, \gW^{(t)}_{ood} \rangle
\end{align}
With Bregman divergence $D_h(\alpha || \alpha_{t-1}) = h(\alpha) - h(\alpha_{t-1}) - \langle \nabla h(\alpha_{t-1}), \alpha - \alpha_{t-1} \rangle$, $h(\alpha) = \sum_i \alpha_i \ln \alpha_i$, we can derive the update rule \ref{apd-equ:update_dw_eval}
\begin{align}
    \alpha_t &:= \argmin_{\alpha \in \Delta^{k}} l_{ood}(\vtheta^{(t+1)}) \notag\\
        &\approx \argmin_{\alpha \in \Delta^{k}} - \eta_t \cdot \langle \alpha, \gW^{(t)}_{ood} \rangle + \mu\cdot D_h(\alpha || \alpha_{t-1}) \notag\\
        &= \argmin_{\alpha \in \Delta^{k}} - \eta_t \cdot\langle \alpha, \gW^{(t)}_{ood} \rangle + \mu (h(\alpha) - \langle \nabla h(\alpha_{t-1}), \alpha\rangle)\notag
\end{align}
\begin{equation}\label{apd-equ:update_dw_eval}
    \Rightarrow \ln \alpha_t^i := \ln \alpha_*^i = \ln \alpha_{t-1}^i + \displaystyle{\frac{\eta_t \gW^{(t)}_{ood}}{\mu}}
\end{equation}


\section{Universal Generalization Evaluation}\label{apdx:universal-eval}
\subsection{Domain Weights on SlimPajama.}
\begin{table*}[htbp!]
\caption{\textbf{Domain weights from \doge with $82$M proxy models.} The results are averaged by three random seeds with standard error.}
\vspace{0.2em}
\label{tab:dw}
\centering
\begin{adjustbox}{max width=0.95\textwidth}
\begin{tabular}{llllllll}
\toprule
 & Arxiv & Book & C4 & CommonCrawl & Github & Stackexchange & Wikipedia\\
\midrule
&0.088\small±0.0008&0.045\small±0.0006&0.269\small±0.0047&0.214\small±0.0101&0.070\small±0.0037&0.166\small±0.0023&0.148\small±0.0061\\
\bottomrule
\end{tabular}
\end{adjustbox}
\end{table*}

\subsection{Evaluation on Various Scales of Base Model.}
We provide the detailed evaluation results on various scale of base model trained on the reweighted pretraining data corpus here. According to the average perplexity, \doge consistently outperforms all the other baseline methods. Besides, \textsc{DoReMi}-$50$k outperforms uniform baseline with both $124$M and $210$M base models, while \textsc{DoReMi}-$10$k fails to get the baseline performance, which suggests \textsc{DoReMi} has a great dependency on the capacity of the auxiliary models. 

\begin{table*}[htbp!]
\caption{\textbf{Perdomain perplexity results on $124/210$M base model.} \doge consistently achieves the best average perplexity over all baseline methods. Performance better than \texttt{Baseline} are in \textbf{Bold}.}
\vspace{0.2em}
\label{tab:124M_ppl}
\centering
\begin{adjustbox}{max width=0.95\textwidth}
\begin{tabular}{lllll||llll}
\toprule
       & & \textbf{$124$M}& & & & \textbf{$210$M} &\\
\midrule
Domain & Baseline & \textsc{DoReMi}-$10k$ & \textsc{DoReMi}-$50k$ & \textsc{DoGE} & Baseline & \textsc{DoReMi}-$10k$ & \textsc{DoReMi}-$50k$ & \textsc{DoGE}\\
\midrule
Arxiv     &  \textbf{8.672}  & 9.353 & 10.149 & 8.954 &  \textbf{8.247}  & 9.041 & 9.637 & 8.456\\
Book   &   51.535  & 57.685 & \textbf{49.038} & 51.564 &   47.060  & 54.393 & \textbf{45.192} & \textbf{46.940}\\
C4 &  56.424 & 63.968 & \textbf{48.494} & \textbf{49.937} &  51.862 & 60.781 & \textbf{44.799} & \textbf{45.588}\\
CommonCrawl & 52.661 & 53.347 & \textbf{47.898} & \textbf{47.297} & 48.319 & 50.456 & \textbf{44.176} & \textbf{43.193}\\
Github &  4.266 & \textbf{4.008} & 4.770 & 4.533  &  4.032 & \textbf{3.871} & 4.510 & 4.234\\
Stackexchange &  9.555 & 9.898 & 10.392 & \textbf{9.365} &  8.948 & 9.477 & 9.760 & \textbf{8.723}\\
Wikipedia & 14.043 & \textbf{13.208} & \textbf{12.246} & \textbf{13.567} &  12.784 & \textbf{12.351} & \textbf{11.358} & \textbf{12.348}\\
\midrule
Average &  18.566 & 19.208 & 18.355 & \textbf{18.066} &  17.218 & 18.285 & 17.119 & \textbf{16.661}\\
\bottomrule
\end{tabular}
\end{adjustbox}
\end{table*}

\begin{figure*}[ht!]
    \centering
    \includegraphics[width=\textwidth]{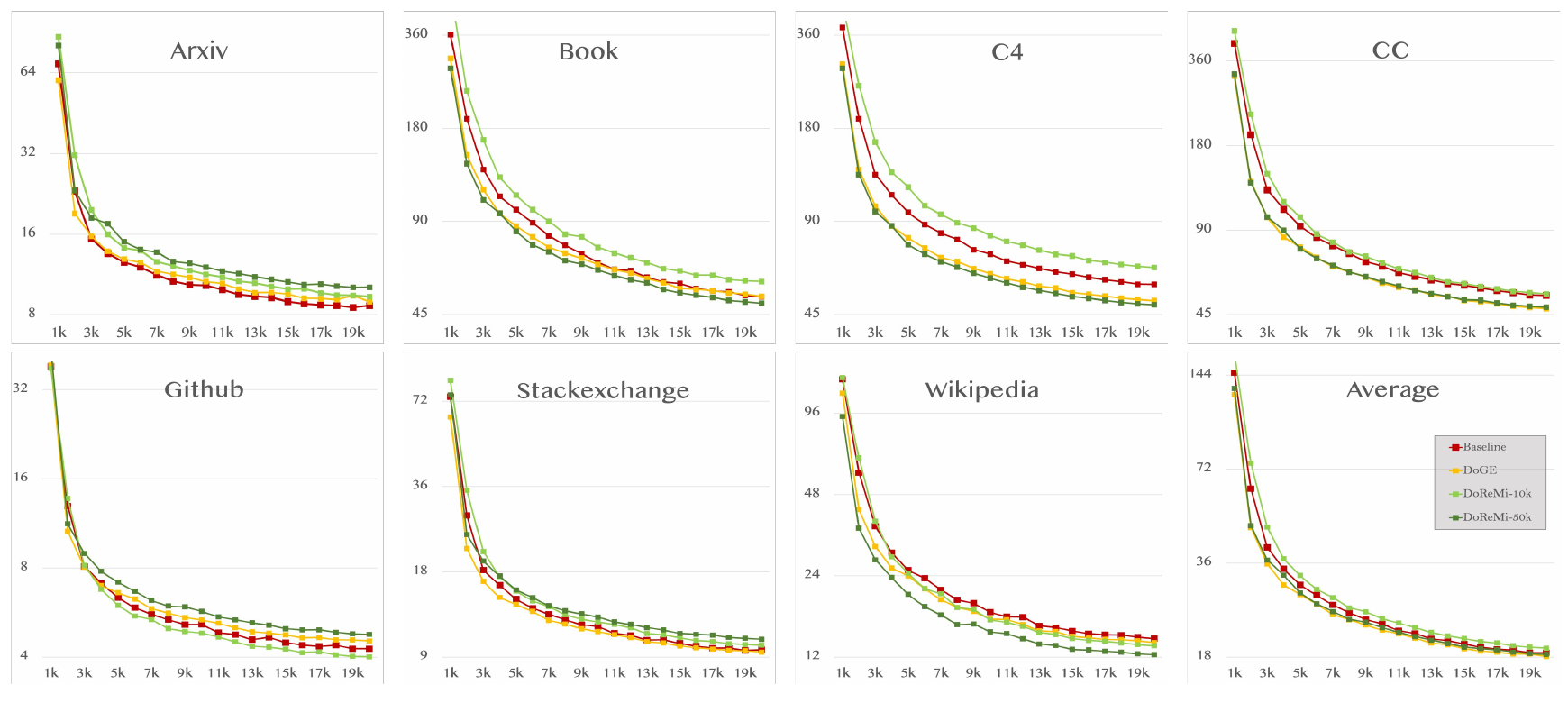}
    \vspace{-1em}
    \caption{\textbf{Perdomain perplexity of 124M base Model.}} 
    \label{fig:124M-ppl}
\end{figure*}

\begin{figure*}[ht!]
    \centering
    \includegraphics[width=\textwidth]{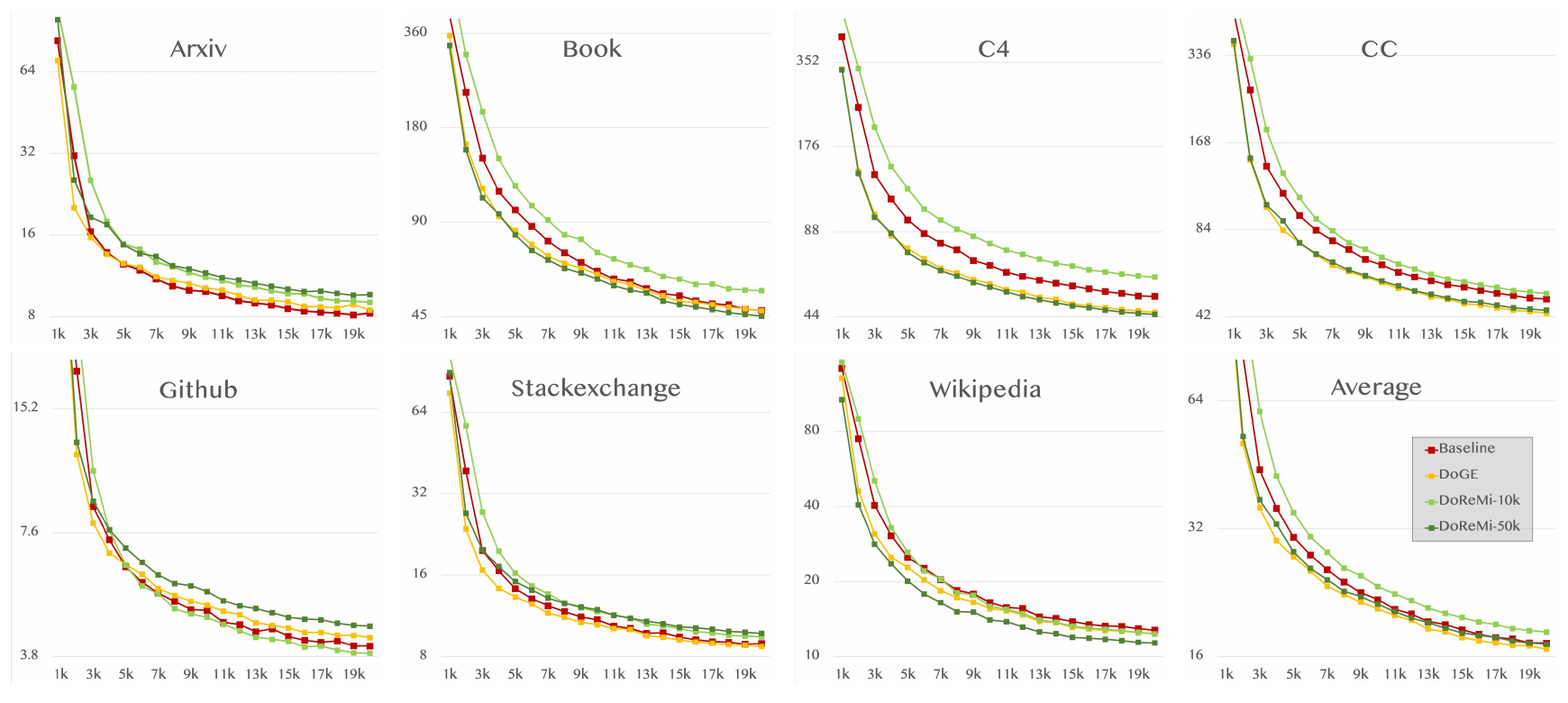}
    \vspace{-1em}
    \caption{\textbf{Perdomain perplexity of 210M base Model.}} 
    \label{fig:210M-ppl}
\end{figure*}

\subsection{Early-stage Training Acceleration.}
\begin{figure*}[ht!]
    \centering
    \includegraphics[width=\textwidth]{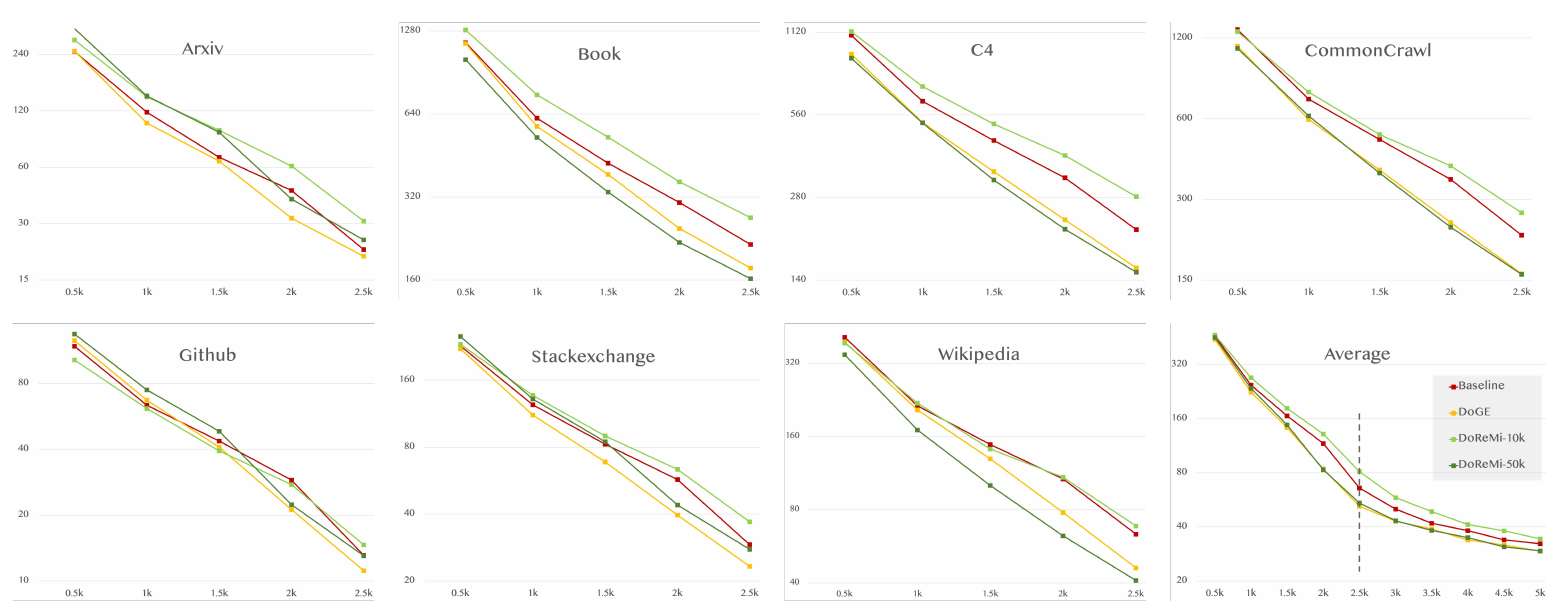}
    \vspace{-1em}
    \caption{\textbf{Perdomain perplexity in early-stage of training (first 2500 steps).} \textsc{DoGE} outperforms baseline across all the domains no matter it is up-weighed or down-weighed.} 
    \label{fig:early_stage}
\end{figure*}

We have observed that \doge reweighted pretraining corpus is able to accelerate the learning process, especially in the early training stage. Fig. (\ref{fig:early_stage}) zooms in into the first 2500 training steps of the base model, where the validation perplexity from \doge drops faster than all the other baseline models on each of all 7 domains, including those are downweighed with less tokens seen. It indicates \doge facilitates the learning of general knowledge, which is shared across domains. 

\subsection{Ablation on the Scale of Proxy Model.}\label{apdx:robustness}
To examine how robust \doge is with various scale of the proxy model, we train \doge with three model scales: $60$M, $82$M and $124$M. All proxy models are trained on the same dataset (Slimpajama) by 10k steps, with the same training hyperparameters. 
Notably, three proxy models with various scales results in consistent domain weights, with only 1.45\% and 0.04\% MAE across 7 domains. Since the difference between three sets of domain weights are negligible, we did not re-train the base model.

\begin{table*}[htbp!]
\caption{\textbf{Domain weights from \doge with various scale of proxy models.} The results are consistent among three different scale of proxy models.}
\vspace{0.2em}
\label{tab:robust}
\centering
\begin{adjustbox}{max width=0.95\textwidth}
\begin{tabular}{llll||lll}
\toprule
Domain & \textsc{DoGE} ($60$M) & \textsc{DoGE}($82$M) & \textsc{DoGE}($124$M)& \textsc{DoReMi} ($60$M) & \textsc{DoReMi} ($82$M) & \textsc{DoReMi} ($124$M)\\
\midrule
Arxiv     &  0.0997  & 0.0880 & 0.0890 &  0.0781  & 0.0424 & 0.0434 \\
Book      &  0.0467  & 0.0450 & 0.0456 &  0.0830  & 0.0819 & 0.0546 \\
C4        &  0.2455  & 0.2693 & 0.2789 &  0.1343  & 0.1141 & 0.1127 \\
CommonCrawl &0.2004  & 0.2135 & 0.1968 &  0.2683  & 0.3811 & 0.3781 \\
Github    &  0.0767  & 0.0703 & 0.0714 &  0.1055  & 0.0654 & 0.0753 \\
Stackexchange&0.1968 & 0.1658 & 0.1703 &  0.1157  & 0.0847 & 0.0919 \\
Wikipedia &  0.1342  & 0.1482 & 0.1480 &  0.2150  & 0.2307 & 0.2440 \\
\midrule
MAE from $82$M proxy &  1.45\% & \slash & 0.48\% &  3.66\% & \slash & 0.91\% \\
\midrule
Computation Time (hours)\footnote{All the models are trained using $4\times A100$ Nvidia GPUs.} & 4.5 & 6.0 & 10.5 &  20.5 & 39.0 & 51.5 \\
\bottomrule
\end{tabular}
\end{adjustbox}
\end{table*}

\ignore{\begin{figure}[ht!]
    \centering
    \includegraphics[width=\textwidth]{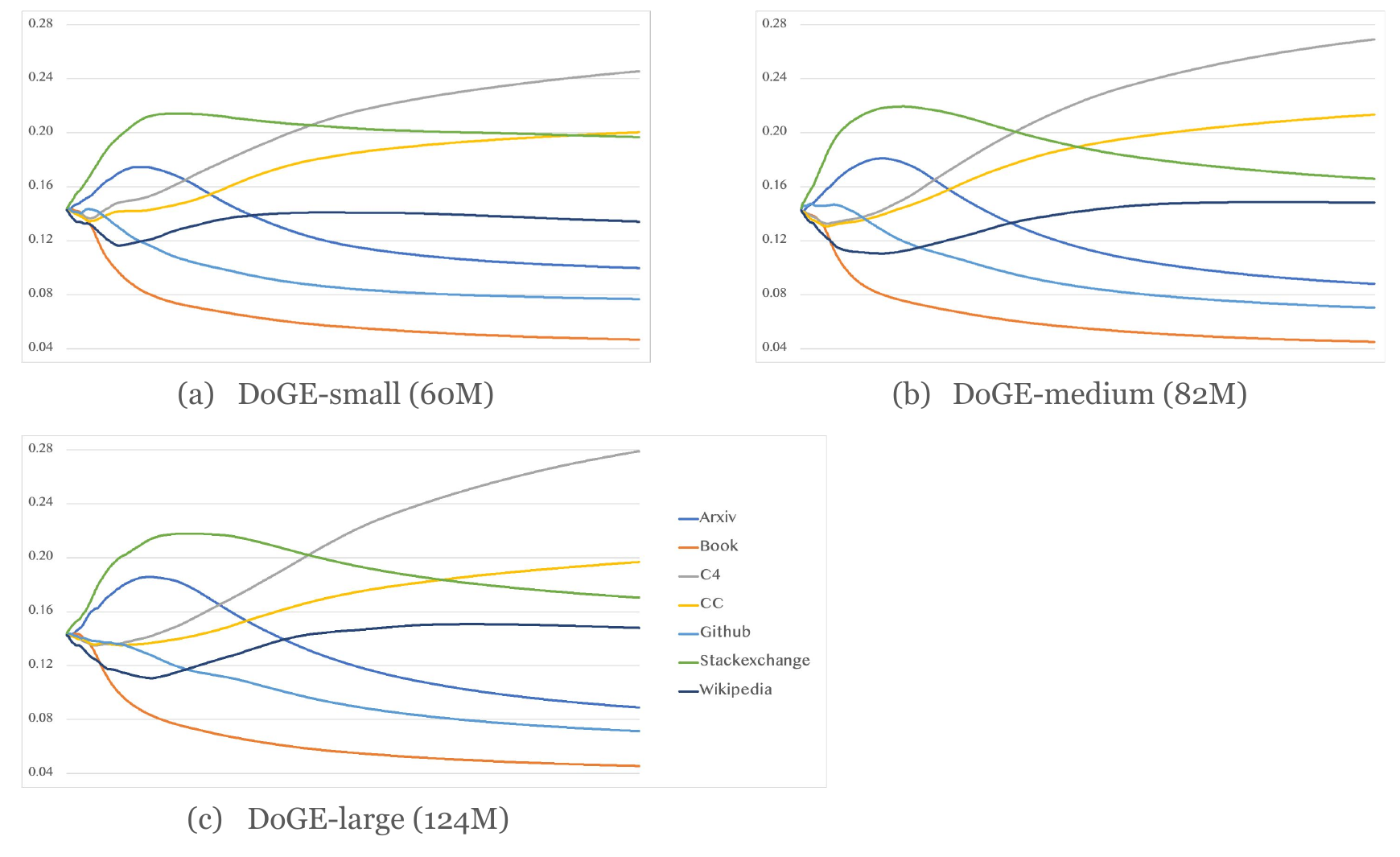}
    \vspace{-1em}
    \caption{\textbf{Domain weights evolution from various scale of proxy models.} \textsc{DoGE} with $60,82,124$M proxy models give consistent domain weights.} 
    \label{fig:robust}
\end{figure}}

\subsection{Performance of Proxy Model.}\label{apdx:proxy}
\begin{figure}[ht!]
    \centering
    \includegraphics[width=\textwidth]{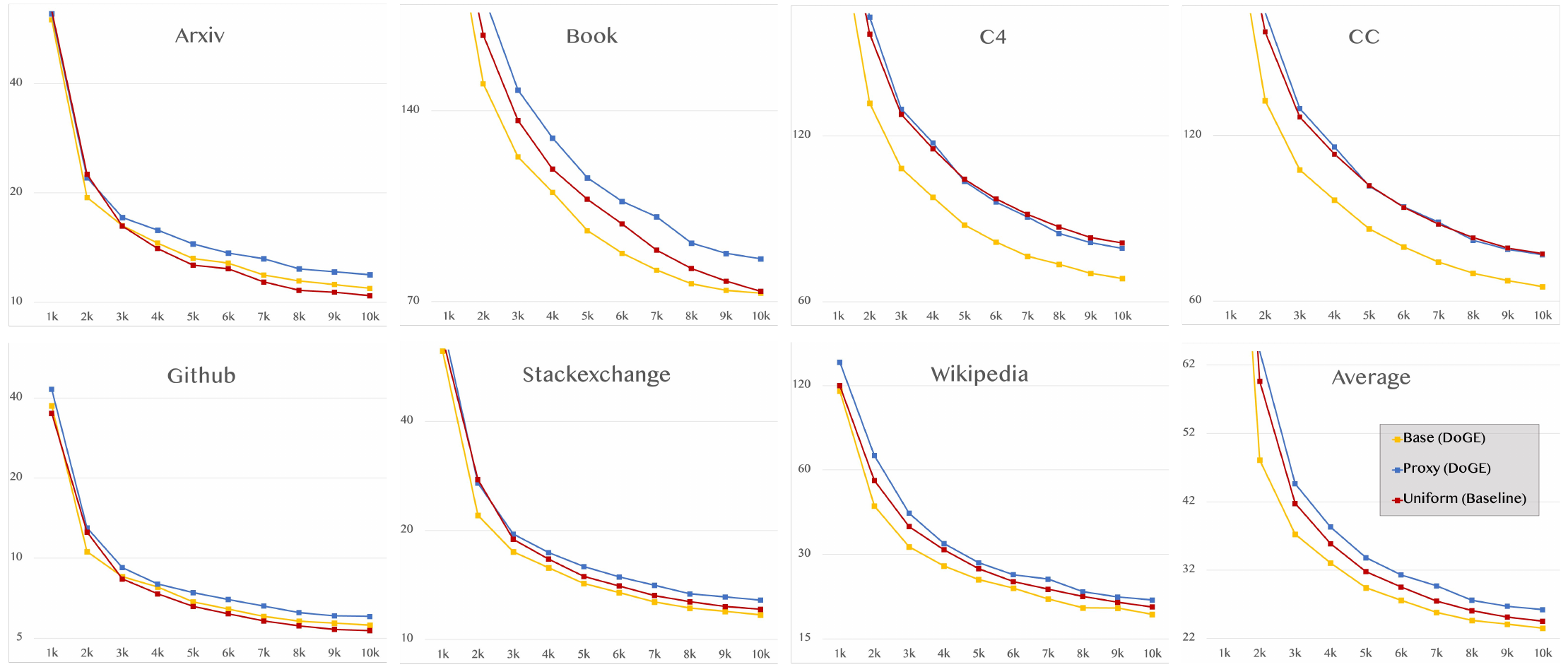}
    \vspace{-1em}
    \caption{\textbf{Comparison of the proxy model and the base model with the same scale ($82$M).} The proxy model shows worse performance on average perplexity than both the base model with resampled training data and uniform sampling weights.} 
    \label{fig:proxy}
\end{figure}

We also compare the performance of the proxy model, which rescales the gradient from each domain at each single step, and the base model trained with resampled training corpus. According to Fig. (\ref{fig:proxy}), the performance of the proxy model falls behind the base model with resampled dataset with \doge domain sampling weights. It is even worse than the baseline with uniform domain weights.

\section{Out-of-Domain Generalization Evaluation}\label{apdx:ood-eval}
\begin{figure*}[ht!]
    \centering
    \includegraphics[width=\textwidth]{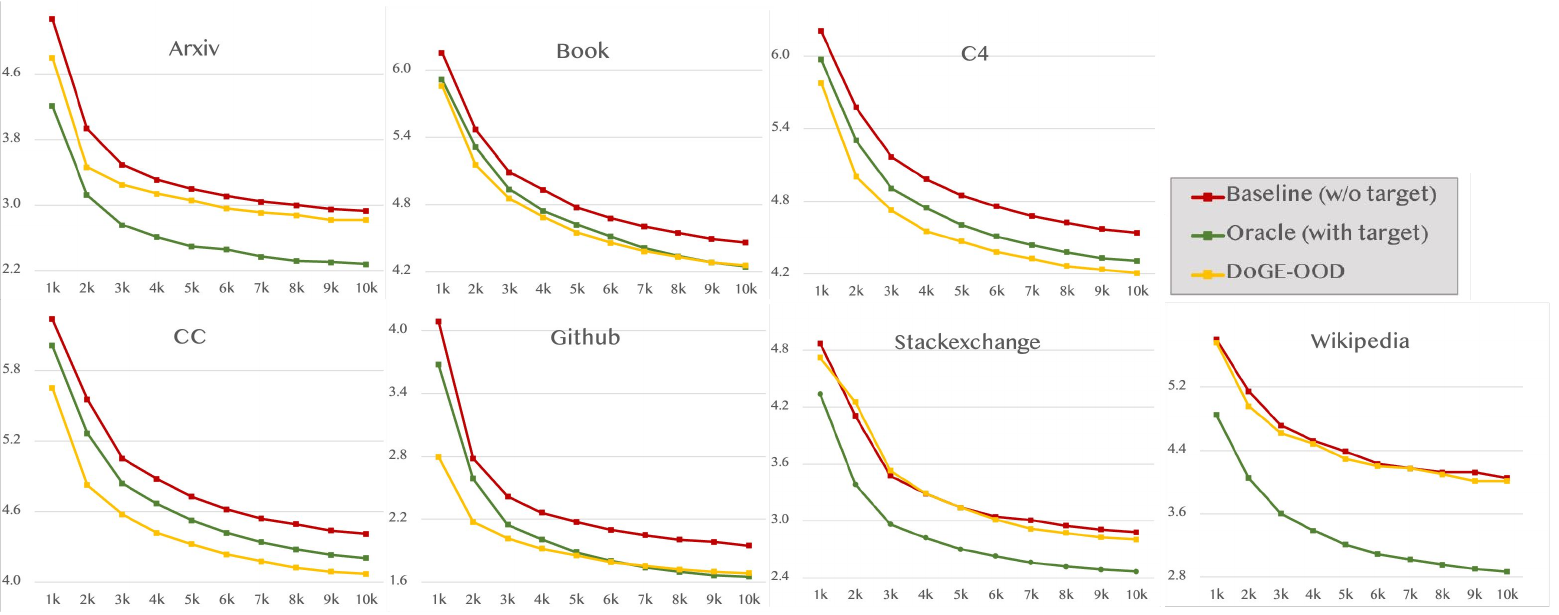}
    \vspace{-1em}
    \caption{\textbf{Target domain perplexity.}} 
    \label{fig:ood_ppl}
\end{figure*}

Fig. \ref{fig:ood_ppl} shows detailed curve of validation perplexity during the training process. On all 7 domains, \doge outperforms uniform baseline without target domain. On Book, Github, C4 and CC, \doge gets comparable or better perplexity than the baseline with access to the target domain. However, on Arxiv, Stackexchange and Wikipedia, both \doge and uniform baseline without target have a large performance gap from the oracle. It indicates learning these domains requires more domain-specific knowledge, which can hardly be obtained from the other source domains. In that case, the gain from source domain reweighting could be limited. 
\subsection{Domain Weights on OoD tasks.}
\begin{table*}[htbp!]
\caption{\textbf{Domain weights from \doge ($82$M) on SlimPajama.} The results are averaged by two random seeds with standard error.}
\vspace{0.2em}
\label{tab:ood-dw}
\centering
\begin{adjustbox}{max width=0.95\textwidth}
\begin{tabular}{llllllll}
\toprule
& & & & Target Domain & & & \\
\midrule
 & Arxiv & Book & C4 & CommonCrawl & Github & Stackexchange & Wikipedia\\
\midrule
Arxiv& 0 &0.063\small±0.0006 &0.035\small±0.0002 &0.045\small±0.0017&0.084\small±0.0003&0.095\small±0.0030&0.091\small±0.0201\\
Book&0.010\small±0.0018 &0&0.117\small±0.0008&0.124\small±0.0071&0.051\small±0.00002&0.048\small±0.0015&0.149\small±0.0102\\
C4&0.125\small±0.0010 &0.341\small±0.0036&0&0.674\small±0.0145&0.058\small±0.0008&0.099\small±0.0045&0.201\small±0.0056\\
CommonCrawl&0.142\small±0.0009 &0.345\small±0.0047&0.721\small±0.0025&0&0.059\small±0.0004&0.076\small±0.0020&0.251\small±0.0258\\
Github&0.161\small±0.0005 &0.055\small±0.0008&0.029\small±0.0002&0.035\small±0.0015&0&0.637\small±0.0124&0.103\small±0.0222\\
Stackexchange&0.366\small±0.0031 &0.081\small±0.0019&0.063\small±0.0008&0.059\small±0.0018&0.691\small±0.0019&0&0.204\small±0.0221\\
Wikipedia&0.106\small±0.0019 &0.113\small±0.0006&0.035\small±0.0004&0.063\small±0.0023&0.057\small±0.0003&0.045\small±0.0013&0\\
\bottomrule
\end{tabular}
\end{adjustbox}
\end{table*}

\begin{table*}[htbp!]
\caption{\textbf{Domain weights from \doge ($82$M) on Wiki40B.} The results are averaged by two random seeds with standard error.}
\vspace{0.2em}
\label{tab:ood-wiki-dw}
\centering
\begin{adjustbox}{max width=0.95\textwidth}
\begin{tabular}{llllll}
\toprule
 & English (en) & German (de) & French (fr) & Spanish (es) & Russian (ru) \\
\midrule
Catalan (ca)& 0.073\small±0.008 &0.043\small±0.008 &0.344\small±0.103&0.516\small±0.102&0.024\small±0.0001\\
Dutch (nl)&0.259\small±0.003 &0.267\small±0.037&0.176\small±0.043&0.203\small±0.025&0.095\small±0.022\\
\bottomrule
\end{tabular}
\end{adjustbox}
\end{table*}

\section{Stage-wise Curriculum}\label{apdx:curriculum}
\begin{figure}[ht!]
  \centering
  \begin{subfigure}[$K=2$]{
  \includegraphics[width=0.6\linewidth,clip]{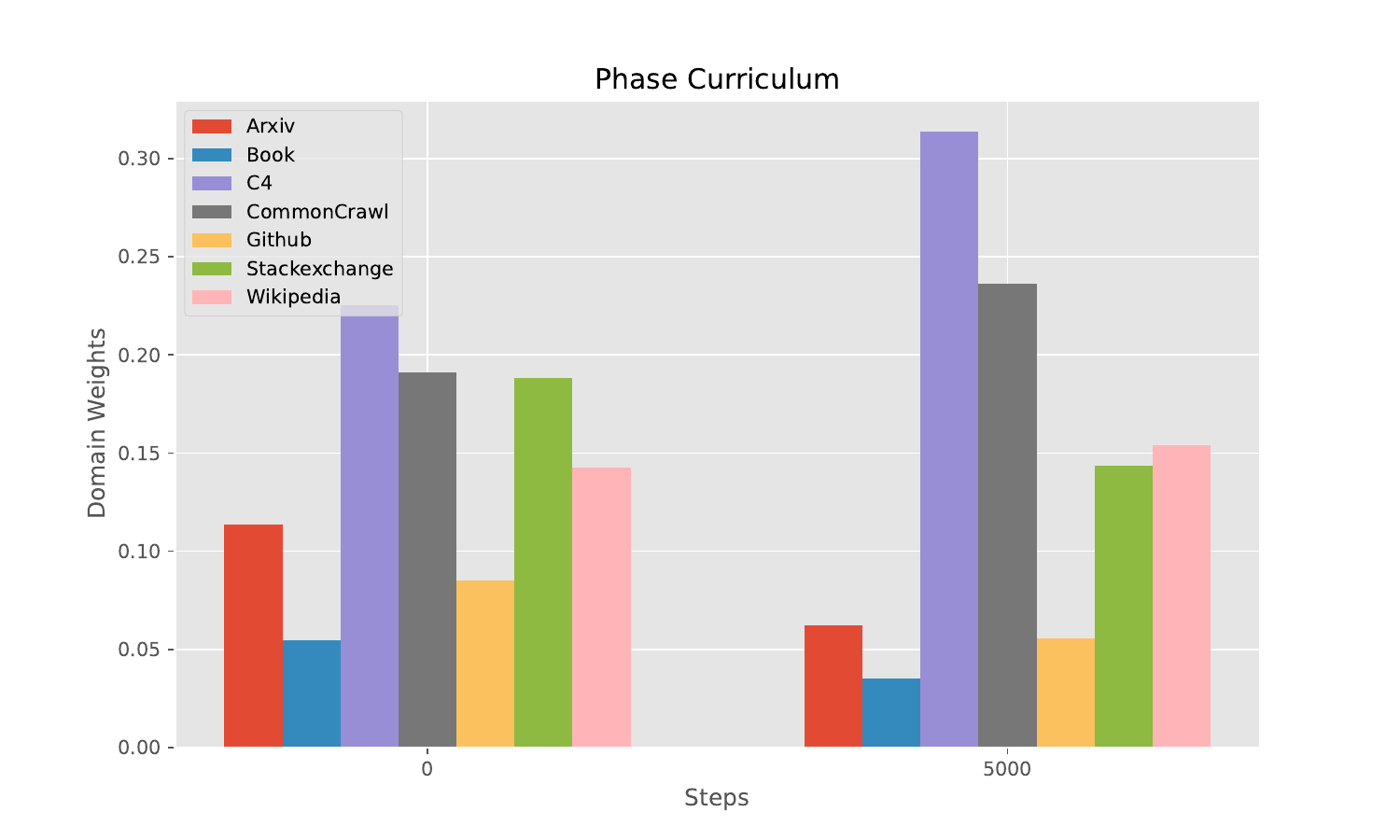} 
  } 
  \end{subfigure} 
  \begin{subfigure}[$K=3$]{
  \includegraphics[width=0.6\linewidth,clip]{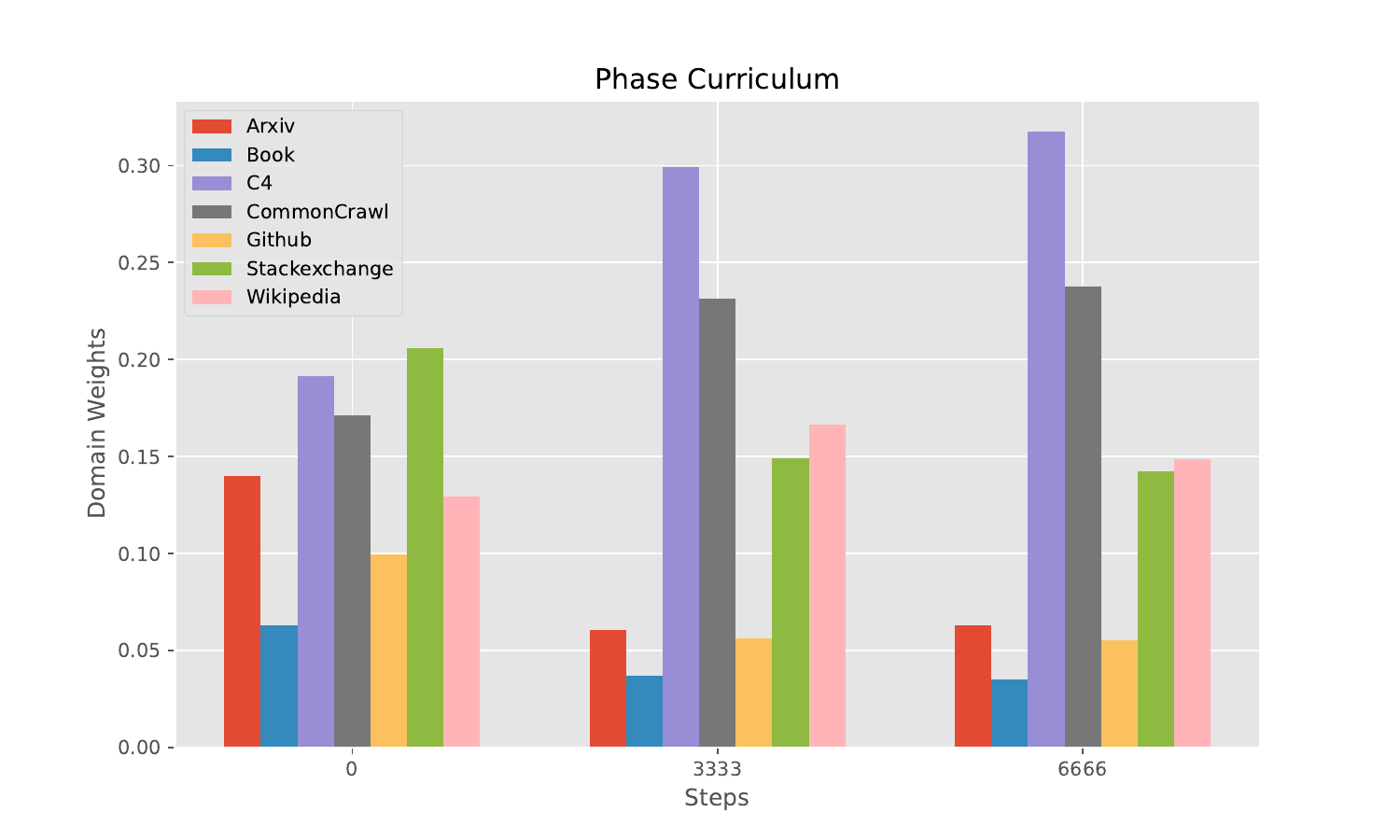} 
  } 
  \end{subfigure} 
  \begin{subfigure}[$K=10$]{
  \includegraphics[width=0.6\linewidth,clip]{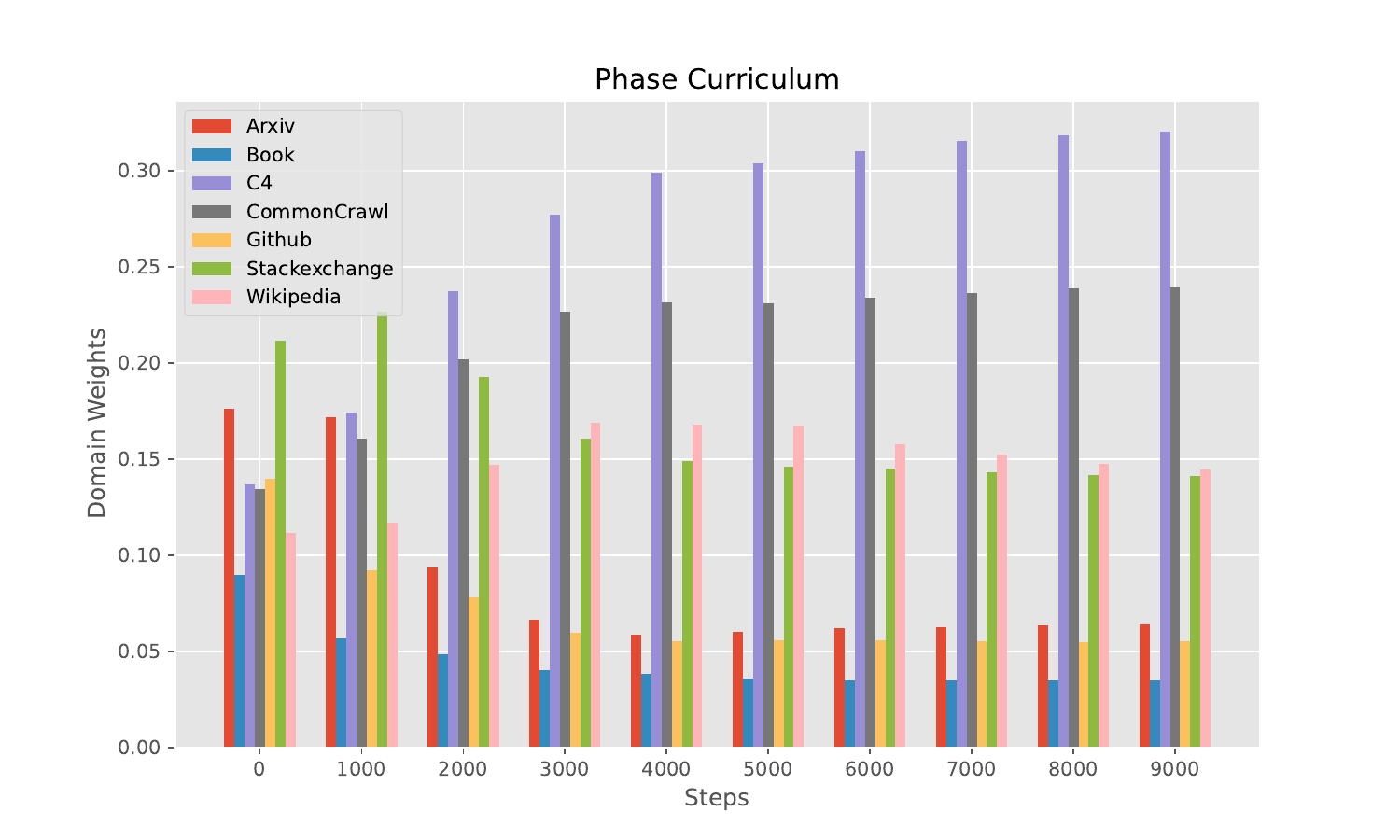} 
  } 
  \end{subfigure} 
 \caption{\textbf{Stage-wise curriculum with $K=2,3,10$ learning stages.} }
 \label{fig:curriculum_dw}
\end{figure}

We provide the implementation and evaluation details of stage-wise curriculum learning in this section. Specifically, we firstly train a $82$M proxy model applying \doge for 10k steps. We then divide the whole training process of the proxy model into $K=2,3,10$ stages, with $5000, 3333, 1000$ training steps in each stage. 
By average the domain weights by number of steps within each stage, we get the stage-wise sampling weights distribution as Fig. (\ref{fig:curriculum_dw}). 
We then train another $124$M model from scratch for 10k steps, where we map the stage-wise sampling weights within the corresponding training steps. We compare the validation perplexity between the model trained with stage-wise curriculum and applying a globally-averaged domain weights. The models trained by each curriculum should have seen the same amount of tokens from each domains in expectation.

With $K=2,3$, the stage-wise curriculum keeps comparable performance as original \doge, which applies the global average as the sampling weights throughout the whole training process. 
It is worth noting that the models learns \emph{hard} domains (C4, CC, Book) slightly better than the global curriculum, while sacrificing the performance on easier domains (Arxiv, Github). 
However, with an extremely find-grained curriculum ($K=10$), the curriculum severely hurt the performance on all the domains by a large margin. 
It suggests that given the same set of data covering diverse knowledge fields, the order of training data does impact the language modelling effectiveness, so that we have to carefully determine the granularity of the curriculum. 

\begin{figure*}[ht!]
    \centering
    \includegraphics[width=\textwidth]{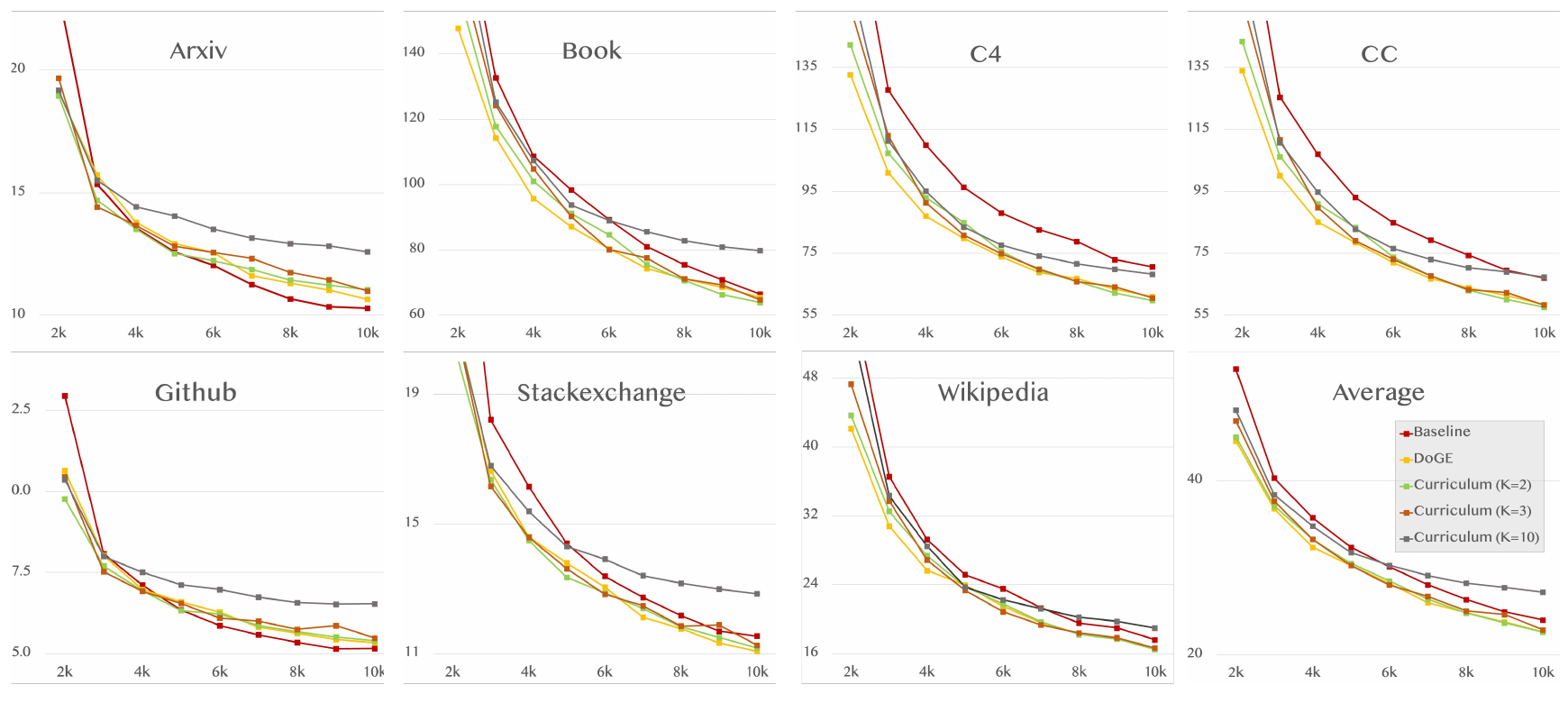}
    \vspace{-1em}
    \caption{\textbf{Per-domain perplexity on validation set with stage-wise curriculum.}} 
    \label{fig:curriculum_ppl}
\end{figure*}

\section{Cancellation Effect}\label{apdx:cancellation}
Following \citet{yeh2022better}, at each time step $t$, we measure the ratio of the actual weight change and the summation of gradient among the mini-batch $B^{(t)}$ for each module of model weights $\vw$. We sum up the ratio across the first $T_c=1000$ steps to obtain the score of cancellation effect $C(\vw)$ as:
\begin{equation}\label{equ:cancellation}
    C(\vw)=\sum_{t \in [T_c]} \frac{\|\vw^{(t+1)}-\vw^{(t)}\|}{\sum_{(x_i\in B^{t})}\frac{\partial l(x_i)}{\partial \vw}}
\end{equation}

During the measurement of cancellation effect, the mini-batch is sampled uniformly from all domains. After the first 1000 steps, we re-initialize the proxy model and compute gradient estimation $\gW$ only using the gradient of the selected parameter modules. 

We then rank all $76$ modules from the parameters of the $82$M proxy model, and apply five parameter selection strategies: (1) We select $K=10,30,50$ modules with the \emph{lowest cancellation effect} scores, denoting \textsc{DoGE}-($low10$,$low30$,$low50$); (2) We select $K=10,30$ modules with the \emph{highest cancellation effect} scores, denoting \textsc{DoGE}-($high10$,$high30$).

\begin{figure*}[ht!]
    \centering
    \includegraphics[width=0.75\textwidth]{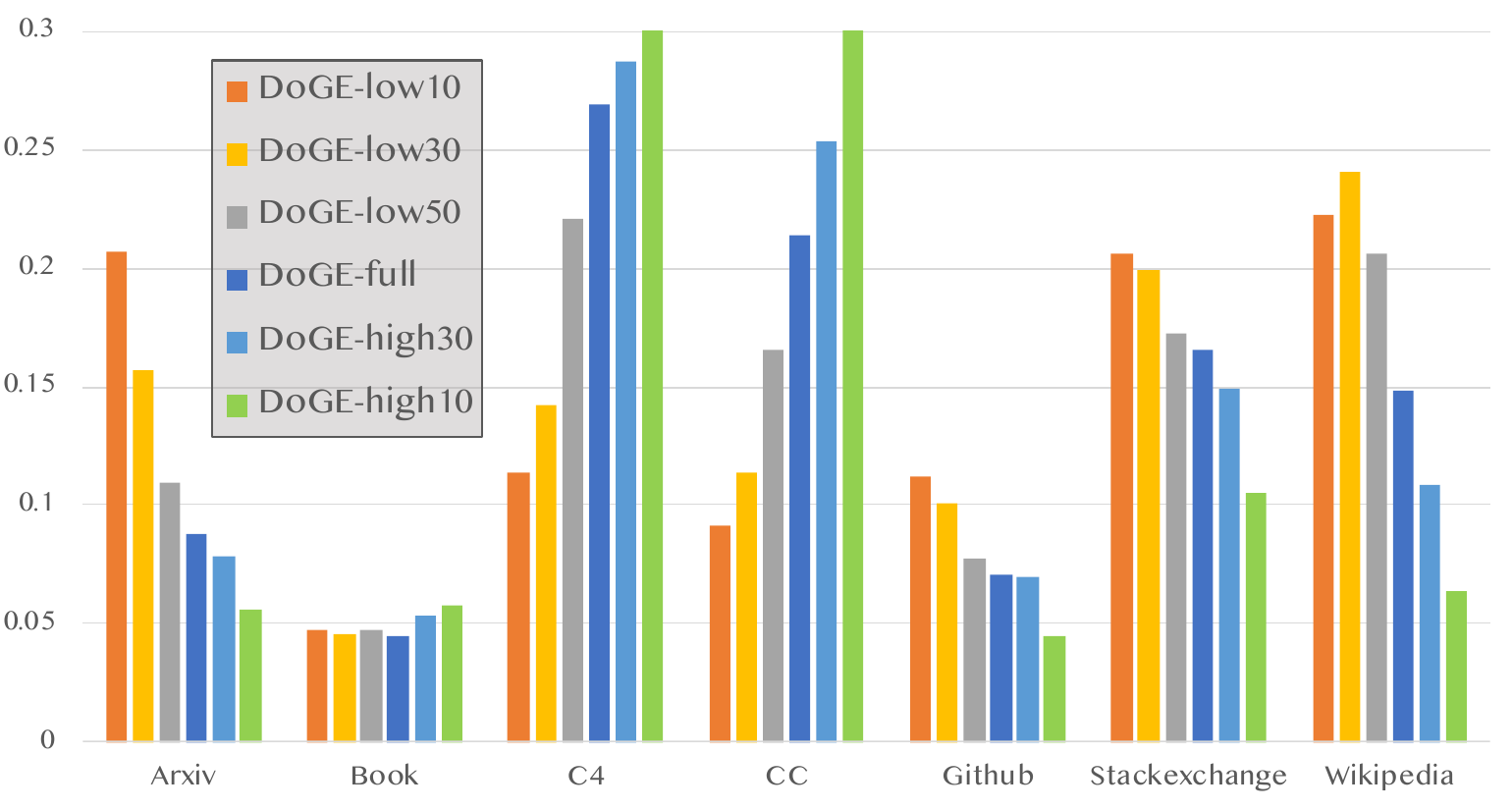}
    \vspace{-1em}
    \caption{\textbf{Domain weights with parameter selection by cancellation effect.}} 
    \label{fig:cc-dw}
\end{figure*}

According to Table. (\ref{tab:cc-ppl}), none of the parameter selection strategies could outperform the original \doge, where the gradient estimation $\gW$ is computed using the full gradient of the proxy model. However, the domain weights from different parameter selection strategies shows an intriguing pattern (Fig. \ref{fig:cc-dw}): modules with low cancellation effect incline to upweigh \emph{unique} domains, which contain more domain specific knowledge (e.g. Wikipedia, Arxiv, Stackexchange), while modules with high cancellation effect tend to upweigh \emph{diverse} domains, which have broader knowledge coverage (e.g. CC, C4). It aligns with the out-of-domain generalization experiment (\S~\ref{sec:ood-exp}), where Wikipedia and Stackexchange get least improvement from domain reweighting, which indicates the \emph{uniqueness} of the domain-specific knowledge.

\begin{table*}[htbp!]
\caption{\textbf{Domain weights with parameter selection based on cancellation effect.} \textsc{DoGE}-low[$k$] (\textsc{DoGE}-high[$k$]) denotes the $k$ modules with lowest (highest) cancellation effect are selected to compute $\gW$.}
\vspace{0.2em}
\label{tab:cc-dw}
\centering
\begin{adjustbox}{max width=0.95\textwidth}
\begin{tabular}{ll||lll||ll}
\toprule
Domain & \textsc{DoGE}-\emph{full} & \textsc{DoGE}-\emph{low10} & \textsc{DoGE}-\emph{low30} & \textsc{DoGE}-\emph{low50} & \textsc{DoGE}-\emph{high30} & \textsc{DoGE}-\emph{high10} \\
\midrule
Arxiv     & 0.08800 &  0.2071  & 0.1571 & 0.1094  & 0.07855 & 0.05635\\
Book      &0.04500 &  0.04734  & 0.04601 & 0.04708  & 0.05304 & 0.05783\\
C4        &0.2693 &  0.1139  & 0.1425 & 0.2209  & 0.2871 & 0.3406\\
CommonCrawl &0.2135 &0.09111  & 0.1142 & 0.1658  & 0.2537 & 0.3316\\
Github    &0.07027 &  0.1123  & 0.1005 & 0.07782 & 0.06943 & 0.04462\\
Stackexchange&0.1658 &0.2061 & 0.1994 & 0.1726 & 0.1494 & 0.1050\\
Wikipedia & 0.1482 &  0.2221  & 0.2402 & 0.2063  & 0.1088 & 0.06402\\
\bottomrule
\end{tabular}
\end{adjustbox}
\end{table*}

\begin{table*}[htbp!]
\caption{\textbf{Validation perplexity by domains with parameter selection based on cancellation effect.} \textsc{DoGE}-low[$k$] (\textsc{DoGE}-high[$k$]) denotes the $k$ modules with lowest (highest) cancellation effect are selected to compute $\gW$.}
\vspace{0.2em}
\label{tab:cc-ppl}
\centering
\begin{adjustbox}{max width=0.95\textwidth}
\begin{tabular}{ll||llll||ll}
\toprule
Domain & Baseline (Uniform) & \textsc{DoGE}-\emph{low10} & \textsc{DoGE}-\emph{low30} & \textsc{DoGE}-\emph{low50} & \textsc{DoGE}-\emph{full}& \textsc{DoGE}-\emph{high30} & \textsc{DoGE}-\emph{high10} \\
\midrule
Arxiv     &8.672 &  \textbf{8.106}  & 8.447 & 8.735 & 8.954 & 9.092 & 9.413\\
Book      &51.535 &  59.359  & 56.146 & 52.700 &51.564 & 50.030 & \textbf{48.526}\\
C4        &56.424 &  61.225  & 57.866 & 53.652 &49.937 & 48.422 & \textbf{45.789}\\
CommonCrawl &52.661 & 57.891  & 54.500 & 49.487 &47.297 & 45.696 & \textbf{42.990}\\
Github    &\textbf{4.266} & 4.268 & 4.298 & 4.460 &4.533 & 4.543 & 4.796\\
Stackexchange &9.555 & \textbf{9.075} & 9.102 & 9.336 &9.365 & 9.494 & 9.982\\
Wikipedia &14.043 & 12.793 & \textbf{12.399} & 12.755 &13.567 & 14.471 & 16.334\\
\midrule
Average &18.566 & 18.848 & 18.442 & 18.151 & \textbf{18.066}&18.067&18.359\\
\midrule
Computation saved for $\gW$ & \slash & 66.9\% & 42.8\% & 24.4\% & \slash & 97.5\% & 99.9\%\\
\bottomrule
\end{tabular}
\end{adjustbox}
\end{table*}
\begin{figure*}[ht!]
    \centering
    \includegraphics[width=0.75\textwidth]{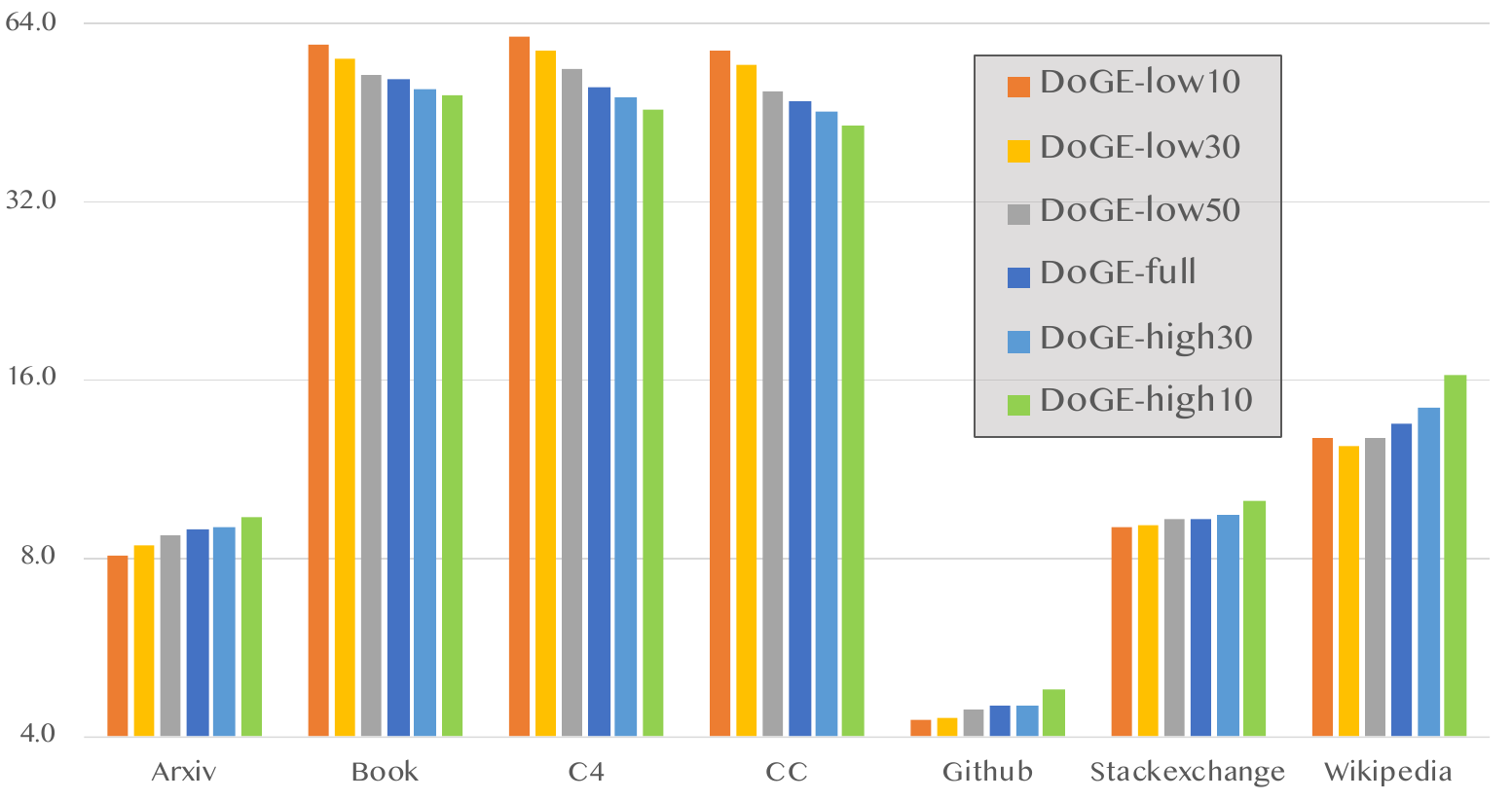}
    \vspace{-1em}
    \caption{\textbf{Validation perplexity by domain with parameter selection by cancellation effect.} } 
    \label{fig:cc-ppl}
\end{figure*}



\end{document}